%% file: main.tex
\pdfoutput=1
\documentclass[12pt]{article}

\usepackage{acro}
\usepackage{amsmath}
\usepackage{amssymb}
\usepackage{amsthm}
\usepackage{bm}
\usepackage{hyperref}
\usepackage[sort,compress]{cleveref}
\usepackage{fullpage}
\usepackage{natbib}
\usepackage{url}
\usepackage{xcolor} 
\usepackage{subcaption}
\usepackage[section]{placeins}

\usepackage{float}
\usepackage{graphicx}
\usepackage{booktabs}
\usepackage[normalem]{ulem}

\newtheorem{proposition}{Proposition}
\newtheorem{lemma}{Lemma}
\newtheorem{definition}{Definition}

\newtheorem{remark}{Remark}

\DeclareMathOperator*{\argmin}{arg\,min}

\DeclareAcronym{PrO}{short=PrO, long=predictively-oriented}
\DeclareAcronym{MCMC}{short=MCMC, long=Markov chain Monte Carlo}
\DeclareAcronym{MFLD}{short=MFLD, long=mean field Langevin dynamics}
\DeclareAcronym{KL}{short=KL, long=Kullback--Leibler}
\DeclareAcronym{LLM}{short=LLM, long=large language model}
\DeclareAcronym{NLL}{short=NLL, long=negative log likelihood}
\DeclareAcronym{SDE}{short=SDE, long=stochastic differential equation}
\DeclareAcronym{LoRA}{short=LoRA, long=low-rank adapter}
\DeclareAcronym{ESS}{short=ESS, long=effective sample size}

\begin{document}

\title{Improving the Optimisation Landscape for Particle Mixtures}

\author{Yan Lin$^1$, David T. Frazier$^2$, Paula Cordero Encinar$^3$ \\
Chris J. Oates$^{1,}$\thanks{Corresponding author.} \\
\small $^1$ Newcastle University, UK \\
\small $^2$ Monash University, Australia \\
\small $^3$ Imperial College London, UK
}

\maketitle

\begin{abstract}
Several methods in statistics and machine learning represent a predictive distribution as an equally weighted mixture of models, or \emph{particles}, trained via the mixture predictive likelihood.
However, we prove that the mixture predictive likelihood is poorly suited to gradient-based optimisation, because it does not identify how the mixture is decomposed across particles, so that a poorly matched particle can be starved of training signal.
As a remedy, we study a convex combination between the \emph{individual risk}, i.e. the average log loss of the individual particles, and the \emph{mixture risk}.
At the particle level, this regularised risk imposes a reverse-KL barrier on the \emph{responsibilities} of the particles, so that no particle can be starved of training signal.
These phenomena are illustrated in two case studies:
In the first, a predictively-oriented posterior is computed using mean-field Langevin dynamics, and the numerics are well-behaved only when our regularisation is employed.
In the second, ensembles of low-rank adapters are jointly fine-tuned on large language models, across four backbones and eight multiple-choice reasoning benchmarks; direct minimisation of the mixture risk leads to responsibility starvation, whose severity decreases as the competence of the backbone increases, whereas our regularised estimator attains a lower held-out mixture risk on every backbone considered.
\end{abstract}

\section{Introduction}
\label{sec:intro}

Consider a supervised learning setting where inputs $x$ are associated to labels $y$ and we are given a training dataset $\{(x_i,y_i)\}_{i=1}^n$.
Given a parametric statistical model $p_{\bm{\theta}}(y|x)$ and a distribution $Q$ over the parameters $\bm{\theta} \in \mathbb{R}^d$, several important and emerging applications in statistics and machine learning are based on the \emph{mixture risk}
\begin{align}
    R_{\mathrm{mix}}(Q) = - \frac{1}{n} \sum_{i=1}^n \log \int p_{\bm{\theta}}(y_i | x_i) \; \mathrm{d}Q(\bm{\theta}) .  \label{eq: log PrO}
\end{align}
This paper is motivated in particular by methods which discretise \eqref{eq: log PrO}  over a finite set of particles $\bm{\theta}_j\in\mathbb{R}^d$:
\begin{align}
Q_m = \frac{1}{m}\sum_{j=1}^m \delta_{\bm{\theta}_j} \quad \implies \quad R_{\mathrm{mix}}(Q_m) = - \frac{1}{n} \sum_{i=1}^n \log \left( \frac{1}{m} \sum_{j=1}^m p_{\bm{\theta}_j}(y_i|x_i) \right) .  \label{eq: discretise Rmix}
\end{align}
Two important examples of \eqref{eq: discretise Rmix}, which we will focus on in the sequel, include the computation of \ac{PrO} posteriors using particle-based numerical methods, and the joint training of ensembles in machine learning.
In both settings, gradient-based minimisation based on the mixture risk \eqref{eq: discretise Rmix} has been empirically observed to perform poorly; our aim is to explain why this phenomenon occurs, and to propose and study a simple remedy to fix it.

\paragraph{Motivation 1: Predictively-Oriented Posteriors}

If the dataset is a genuine sample from the statistical model for some true parameter $\bm{\theta}^\dagger$, then Bayesian updating is a principled framework in which $\bm{\theta}^\dagger$ can be estimated.
However, when modelling assumptions are violated the usefulness of the posterior predictive distribution is limited, since in regular settings the posterior concentrates around a single pseudo-true parameter $\bm{\theta}^\star$ on which (incorrect) predictions are based.
Motivated by this limitation, recent work \citep{shen2025prediction,lai2024predictive,mclatchie2025predictively} is inspired by Bayesian ideas but replaces the standard Bayesian objective, in which parameter uncertainty is central, with an alternative (nonlinear) objective in which \emph{predictive} uncertainty is central.
To be more precise, let $\mathcal{P}(\mathbb{R}^d)$ denote the set of probability distributions on $\mathbb{R}^d$, let $R$ be an (empirical) risk functional on $\mathcal{P}(\mathbb{R}^d)$, and consider 
\begin{align}
    \argmin_{Q \in \mathcal{P}(\mathbb{R}^d)} \; J(Q), \qquad \qquad J(Q) = R(Q) + \frac{1}{n} \mathrm{KL}(Q || \Pi)  \label{eq: obj}
\end{align}
where $\mathrm{KL}(\cdot ||\Pi)$ denotes \ac{KL} divergence relative to a `prior' $\Pi \in \mathcal{P}(\mathbb{R}^d)$.
The standard Bayesian posterior solves \eqref{eq: obj} with risk function 
\begin{align}
R_{\mathrm{ind}}(Q) = - \frac{1}{n} \sum_{i=1}^n \int \log p_{\bm{\theta}}(y_i|x_i) \; \mathrm{d}Q(\bm{\theta}) , \label{eq: loss Bayes}
\end{align}
which can be interpreted as a measure of data fit; we call \eqref{eq: loss Bayes} the \emph{individual risk}, since it averages the log loss of the individual models $p_{\bm{\theta}}$ over $Q$.
Instead, the \ac{PrO} approach takes $R$ to be the mixture risk \eqref{eq: log PrO}, and the corresponding solution of \eqref{eq: obj}, denoted $Q_{\mathrm{PrO}}$, is called a \ac{PrO} \emph{posterior} by analogy with the Bayesian framework.
Observe that \eqref{eq: log PrO} differs from $R_{\mathrm{ind}}(Q)$ only in the order in which the integral and the logarithm are performed. 
This difference is not trivial; as a result, the predictive performance of \ac{PrO} posteriors (asymptotically) dominates that of the Bayesian posterior when the model is misspecified, and (asymptotically) matches it if the model is well-specified \citep[Theorem 1 of][]{mclatchie2025predictively}.
Further, because \eqref{eq: log PrO} directly targets \emph{predictive} performance, the level of uncertainty in predictions is automatically asymptotically calibrated under convex recovery of the data-generating model \citep[Theorem 2 of][]{mclatchie2025predictively}.

Computing a \ac{PrO} posterior is, however, more involved than computing a Bayesian posterior; in the latter case the posterior density is known up to a normalising constant and standard techniques such as \ac{MCMC} can be applied.
Since $R_{\mathrm{mix}}$ is nonlinear in $Q$, the first-order optimality condition for \eqref{eq: obj} is instead a fixed-point equation,
\begin{align*}
    \frac{\mathrm{d}Q_{\mathrm{PrO}}}{\mathrm{d}\Pi}(\bm{\theta}) \propto \exp\left\{ - n R_{\mathrm{mix}}'(Q_{\mathrm{PrO}})(\bm{\theta}) \right\} ,
\end{align*}
where $R'(Q) : \mathbb{R}^d \rightarrow \mathbb{R}$ denotes the first variation of a functional $R$ evaluated at $Q$.
Since no explicit unnormalised density representation is available for $Q_{\mathrm{PrO}}$, alternative numerical methods are required.
Principal among these are methods based on gradient flows of \eqref{eq: obj}, in particular the \ac{MFLD}
\begin{align}
\mathrm{d}\bm{\theta}_j^t & =  [(\nabla \log \pi)(\bm{\theta}_j^t) - n \nabla R_{\mathrm{mix}}'(Q_m^t)(\bm{\theta}_j^t)]  \, \mathrm{d}t + \sqrt{2} \, \mathrm{d}\bm{B}_j^t , \qquad Q_m^t = \frac{1}{m} \sum_{j=1}^m \delta_{\bm{\theta}_j^t} \label{eq:intro-mfld}
\end{align}
where $\pi$ is a density for $\Pi$ and each $\bm{B}_j^t$ is an independent standard Brownian motion on $\mathbb{R}^d$.
Under suitable conditions the measure $Q_m^t$ converges to $Q_{\mathrm{PrO}}$ as both $t$ and $m$ are increased \citep{nitanda2025propagation}.
However, major practical difficulties have been reported in this context, in that oftentimes only a small number of particles receive a meaningful signal through the gradient \citep[e.g. Appendix E.2 of][]{mclatchie2025predictively}.
At present these difficulties present a major barrier to practical deployment.
Further, these difficulties are not ameliorated by the use of alternative gradient flow techniques, such as the \emph{variational gradient descent} method
\citep{wang2019nonlinear,chazal2025computable,liu2025detecting}.
Our work reveals that the root of these issues lies in the mixture risk $R_{\mathrm{mix}}(Q_m^t)$ viewed as a function of the particles $\{\bm{\theta}_j^t\}_{j=1}^m$, not in the choice of gradient flow through which computation is performed.

\paragraph{Motivation 2: Joint Training of Ensembles}

Deep ensembles, which average the predictions of independently trained deep neural networks, are among the most effective practical approaches to uncertainty quantification in deep learning \citep{lakshminarayanan2017deep}.
A natural extension is to train the members of an ensemble \emph{jointly}, by minimising the loss of their averaged prediction; for an ensemble with parameters $\{\bm{\theta}_j\}_{j=1}^m$ this is precisely $R_{\mathrm{mix}}(Q_m)$ in \eqref{eq: discretise Rmix}.
Joint training is in principle attractive because it encourages the members to complement one another, and avoids potentially wasteful computation from training two models that are near-identical, but curiously it has repeatedly been observed to underperform independent training \citep{webb2020ensemble,jeffares2023joint}.

The question has become pressing for \acp{LLM} in settings where a calibrated confidence is required.
Here, ensembles of parameter-efficient adapters on a shared frozen backbone can be simultaneously trained \citep{nitanda2025propagation,cordero2026soft}, typically using a noisy variant of AdamW \citep{nitanda2025propagation,wang2026distributional}:
\begin{align}
\bm{\theta}_j^{k+1} = \bm{\theta}_j^k + \eta \, \bm{D}_j^k + \sqrt{\frac{2\eta}{mn}} \, \bm{\xi}_j^k , \qquad V(\bm{\theta}_{1:m}) = \sum_{j=1}^m \log \pi(\bm{\theta}_j) - mn R_{\mathrm{mix}}(Q_m) , \label{eq:intro-noisy-adamw}
\end{align}
where $\eta$ is a learning rate, $\bm{D}_j^k$ is the AdamW direction computed from the gradients $\nabla_{\bm{\theta}_j} V$ at iterations $0,\dots,k$, $\pi$ is a (typically Gaussian) prior density, and each $\bm{\xi}_j^k$ is an independent standard Gaussian vector.
Since $\nabla_{\bm{\theta}_j} V$ is exactly the drift of \ac{MFLD} in \eqref{eq:intro-mfld}, replacing $\eta \bm{D}_j^k$ by the raw gradient step $\frac{\eta}{mn} \nabla_{\bm{\theta}_j} V$ would recover an Euler--Maruyama discretisation of \ac{MFLD} with step size $\eta/(mn)$.
In fact, AdamW rescales the drift by a running estimate of the gradient magnitude while leaving the noise unchanged \citep{nitanda2025propagation}, so the update runs at a much lower effective temperature than \ac{MFLD} and the particles instead approximately maximise $V$; that is, they approximately minimise the mixture risk $R_{\mathrm{mix}}(Q_m)$ plus a small ridge penalty.
However, in practice the benefit of joint training appears not to be realised; \citet{wang2026distributional} reported gains that were at best incremental in the context of fine-tuning \acp{LLM}.
Again, our work reveals that the root of this issue is the mixture risk $R_{\mathrm{mix}}(Q_m)$ viewed as a function of the particles $\{\bm{\theta}_j\}_{j=1}^m$.

\paragraph{Our Contributions}

\begin{figure}[!tbp]
\centering
\includegraphics[width=\textwidth]{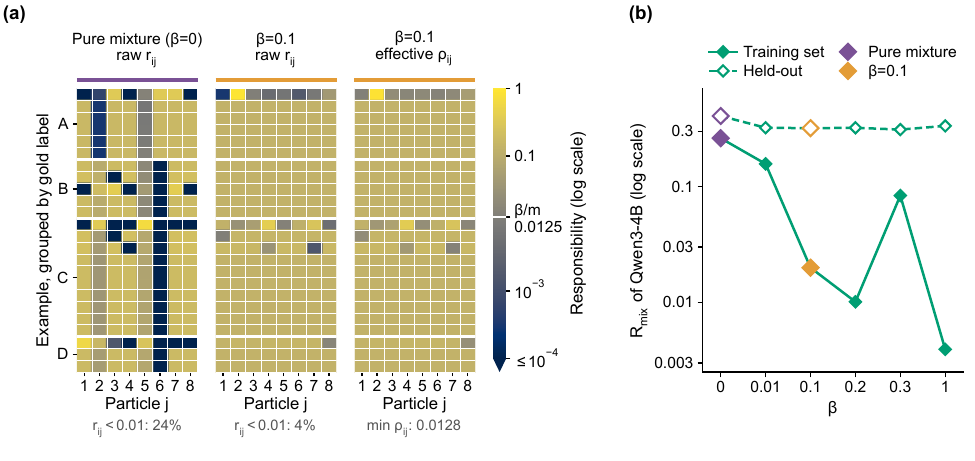}
\caption{Illustration of responsibility starvation for an ensemble of $m=8$ \acp{LLM} jointly fine-tuned using noisy AdamW \eqref{eq:intro-noisy-adamw}. 
(a) Responsibilities $r_{ij}$ of the trained particles of Llama 3.2-3B on 24 HellaSwag test examples, under pure mixture training ($\beta=0$) and the bridge ($\beta=0.1$), together with the \emph{effective responsibilities} $\rho_{ij}$ of the bridge, defined in \Cref{subsec: bayespro}. 
Under pure mixture training, particle 6 has $r_{ij}<0.01$ on 18 of the 24 examples and on $75\%$ of all $10{,}042$ test examples. (b) Training-set and held-out $R_{\mathrm{mix}}$ of Qwen3-4B after three epochs for six values of $\beta$. 
The run with $\beta=0$ attains the largest value of $R_{\mathrm{mix}}$. }
\label{fig:mechanism-puzzle}
\end{figure}

The mixture risk \eqref{eq: log PrO} is convex in $Q$, so in principle solving the optimisation problem in \eqref{eq: obj} should be straightforward.
However, \emph{this convexity is lost when $Q$ is restricted to the particle approximation $Q_m$}, because the map $(\bm{\theta}_1,\dots,\bm{\theta}_m) \mapsto R_{\mathrm{mix}}(Q_m)$ is in general not convex, even though $Q \mapsto R_{\mathrm{mix}}(Q)$ is.
Our contributions are to diagnose the consequences of this fact, to propose a remedy, and to explore its effectiveness in two detailed case studies:
\begin{itemize}
\item \textbf{Diagnosis: responsibility starvation.}
We show that particle-based minimisation of the mixture risk is prone to failure, because the mixture risk constrains the \emph{aggregate} predictive mixture formed by the particles $\{\bm{\theta}_j\}_{j=1}^m$, not how that mixture is \emph{decomposed} across the particle set.
This leaves room for a poorly matched particle to receive a vanishing share of the training signal, meaning that certain data points $(x_i,y_i)$ enter the gradient update for a particle $\bm{\theta}_j$ with vanishing \emph{responsibility} weight $r_{ij}$ (see \Cref{sec:methods}); we call this phenomenon \emph{responsibility starvation}.
We prove that the gradient correcting a wrongly specialised particle vanishes as certain degenerate decompositions (which we call \emph{slot partitions}) are approached (\Cref{sec:slot-theory}), so that direct minimisation of the mixture risk, whether by \ac{MFLD} or by any similar gradient-based method, is predisposed to fail.

\item \textbf{Remedy: the bridge.}
We analyse a simple regularised risk, which we call the \emph{bridge} risk, which interpolates between the mixture risk \eqref{eq: log PrO} (at $\beta=0$) and the individual risk \eqref{eq: loss Bayes} (at $\beta=1$), as a means of repairing the optimisation landscape.
At the particle level, we show that the bridge imposes a reverse-KL barrier on responsibilities $r_{ij}$, so that no particle can be starved and a degenerate decomposition is escaped in finite time (\Cref{sec:methods,sec:slot-theory}); at the distributional level, we show that, for \ac{PrO} posteriors, it is equivalent to regularising the \ac{PrO} posterior toward a $\beta$-tempered Bayesian target (\Cref{sec:case-pro}).

\item \textbf{Case Study~1: computing \ac{PrO} posteriors.}
Through the lens of a simple logistic regression model, we demonstrate that a \ac{PrO} posterior computed using \ac{MFLD} with a small positive amount $0 < \beta \ll 1$ of regularisation substantially out-performs the vanilla use of \ac{MFLD} in approximating $Q_{\mathrm{PrO}}$, despite the mixture risk no longer being the optimisation target (\Cref{sec:case-pro}).
Indeed, here \ac{MFLD} fails to converge, even with $m=128$ particles and after $10^{7}$ iterations. 
On the other hand, with $\beta=0.01$ every run converges to the same solution within about $10^{5}$ iterations using only $m=32$ particles (\Cref{subsec:cs1-numerics}).

\item \textbf{Case Study~2: joint training of \ac{LLM} ensembles.}
When ensembles of \acp{LoRA} are jointly fine-tuned on four \ac{LLM} backbones and eight multiple-choice reasoning benchmarks, pure mixture training exhibits responsibility starvation, and the bridge with a small $\beta$ such as $\beta=0.1$ achieves lower held-out values of the \emph{original} mixture risk on every backbone, substantially so on three of the four, despite the mixture risk no longer being the optimisation target (\Cref{sec:experiments}).
\Cref{fig:mechanism-puzzle} previews these results.
\end{itemize}

\noindent In short, we show that even a small $\beta$ can dramatically improve the optimisation landscape for particle mixtures, and we therefore recommend the use of a small $\beta$ in general.
Our code is available at \url{https://github.com/YanLin-Quinne/bayes-pro-bridge}.

\paragraph{Related Work}

The use of \eqref{eq: log PrO} as a training objective dates back several decades under the name \emph{nonparametric maximum likelihood} \citep[NML, see Chapter 5 of][]{lindsay1995mixture}.
Related predictive objectives have since been studied by \citet{masegosa2020learning,jankowiak2020parametric,sheth2020pseudo,morningstar2022pacm}.
However, the NML estimator has discrete support, which makes its computation difficult or impossible in general.
The regularised formulation \eqref{eq: obj}, introduced concurrently by \citet{shen2025prediction,lai2024predictive}, in principle alleviates computational difficulties by convexifying the objective functional over $\mathcal{P}(\mathbb{R}^d)$.
The same objective arises in the solution of Fredholm integral equations of the first kind, where \citet{crucinio2024solving} minimise an entropically regularised version of the mixture risk over probability measures and approximate the minimiser by the interacting particle system of the associated Wasserstein gradient flow, building on the particle method of \citet{crucinio2023particle}.
However, this convexification breaks down when particle approximation is employed, motivating the present work.

The recent work of \citet{cordero2026soft} proposed a different mechanism to interpolate between the Bayesian and \ac{PrO} posteriors, based on an $\alpha$-R\'enyi risk
\begin{align*}
    R_{\text{R\'{e}nyi}}^\alpha(Q) = - \frac{1}{\alpha n} \sum_{i=1}^n \log \int p_{\bm{\theta}}(y_i|x_i)^\alpha \; \mathrm{d}Q(\bm{\theta})
\end{align*}
where $\alpha \in (0,1]$ is to be specified.
The mixture risk \eqref{eq: log PrO} is recovered by setting $\alpha = 1$, while an application of l'H\^{o}pital's rule reveals that the individual risk \eqref{eq: loss Bayes} is recovered in the limit $\alpha \rightarrow 0$ \citep[Lemma 1 in][]{cordero2026soft}.
The goal of that work was methodological, aiming to balance globally plausible individual models against a system of complementary specialists, whereas our focus is \emph{computational}: we use regularisation to improve the optimisation landscape for the \emph{standard} mixture risk, and the bridge has a clear interpretation as regularising the \ac{PrO} posterior towards a $\beta$-tempered Bayesian target (\Cref{prop:tempered-reference}).

Training of finite mixture models is well known to suffer from under-identifiability of the components, which are identifiable only up to label switching \citep{redner1984mixture,jasra2005markov}; the resulting non-convex, multi-modal landscape causes iterative procedures such as EM to converge to poor local optima, or only very slowly, when components are not well separated or the model is over-parametrised \citep{xu1996convergence,jin2016local,dwivedi2020singularity}.

Minimisation of $R_{\mathrm{mix}}(Q_m)$ is the joint, or end-to-end, training of an ensemble whose members' predicted probabilities are averaged.
\citet{webb2020ensemble} observed that end-to-end training of ensembles can underperform independent training, and studied objectives that interpolate between the two.
\citet{jeffares2023joint} attributed the failure of joint training to \emph{learner collusion}, in which ensemble members artificially inflate their apparent diversity on the training data, so that the joint fit fails to generalise, and proposed an augmented objective that interpolates between independent and joint training.
For the log loss with averaged probabilities, their augmented objective coincides with the bridge $R_\beta(Q_m)$ of \Cref{def:bayes-pro-bridge}.
The mechanism identified here is different: it is a failure of \emph{optimisation} rather than of generalisation, since direct minimisation of $R_{\mathrm{mix}}$ attains a worse value of $R_{\mathrm{mix}}$ than the bridge on the training data itself (\Cref{fig:mechanism-puzzle}), which we explain through the non-identifiability of the decomposition of the mixture across particles (\Cref{sec:slot-theory}).

A related difficulty arises in multiple choice learning, where each example trains only the best of several hypotheses, so that some hypotheses can receive no gradient \citep{guzman2012multiple,lee2016stochastic}; the \emph{relaxed} winner-takes-all rule of \citet{rupprecht2017learning} assigns weight $1-\varepsilon$ to the best hypothesis and $\varepsilon/(m-1)$ to each of the other $m-1$, a structure closely mirrored by the \emph{effective responsibilities} of the bridge we discuss in \Cref{subsec: bayespro}.
Responsibility-weighted gradients of this kind also arise in the training of mixtures of experts \citep{jacobs1991adaptive}.

\section{Methods}
\label{sec:methods}

This section introduces the mathematical formulation of \emph{responsibility} in \Cref{subsec:particle-mixtures}, explaining how vanishing responsibility leads to a loss of training signal, and then explains our simple remedy to improve the optimisation landscape for particle mixtures in \Cref{subsec: bayespro}.
The simplicity of our methods is viewed as a major strength; the main technical effort is devoted to establishing their effectiveness theoretically (\Cref{sec:slot-theory}) and empirically (\Cref{sec: experiments}).

\subsection{Responsibilities and Loss of Training Signal}
\label{subsec:particle-mixtures}

In this section we suppose that particles $\{\bm{\theta}_j\}_{j=1}^m$ define an empirical measure $Q_m$ at which the mixture risk \eqref{eq: log PrO} is evaluated; that is, we are in the setting of \eqref{eq: discretise Rmix}.
To explain the fundamental issue with particle-based optimisation of the mixture risk, we introduce the concept of \emph{responsibility}:

\begin{definition}[Responsibility]
The \emph{responsibility} of particle $j$ for example $i$ is defined as
    $$
    r_{ij}=\frac{p_{\bm{\theta}_j}(y_i\mid x_i)}{\sum_{k=1}^m p_{\bm{\theta}_k}(y_i\mid x_i)} .
    $$
\end{definition}

\noindent To interpret the responsibility through a Bayesian lens, $r_{ij}$ is the posterior probability that datum $y_i$ was generated by the model $p_{\bm{\theta}_j}(\cdot | x_i)$ given a uniform prior distribution over the set of all models $\{p_{\bm{\theta}_k}(\cdot | x_i)\}_{k=1}^m$.

The motivation for considering responsibilities comes from considering the update of particle $\bm{\theta}_j$ as driven by the gradient of the mixture risk:
\begin{equation}
\nabla_{\bm{\theta}_j} R_{\mathrm{mix}}(Q_m) = -\frac{1}{n}\sum_{i=1}^n r_{ij}\, \nabla_{\bm{\theta}} \log p_{\bm{\theta}_j}(y_i\mid x_i) 
\label{eq:mix-gradient}
\end{equation}
Thus the responsibility determines how strongly example $i$ contributes to the update of particle $j$, and we refer to it informally as the \emph{training signal} that particle $j$ receives from example $i$.
By contrast, 
\begin{align*}
\nabla_{\bm{\theta}_j} R_{\mathrm{ind}}(Q_m) = -\frac{1}{mn}\sum_{i=1}^n \nabla_{\bm{\theta}} \log p_{\bm{\theta}_j}(y_i\mid x_i) ,
\end{align*} 
in which every example contributes to the update of every particle with an equal amount of training signal.
Particles which receive little or no responsibility for any of the examples in the training dataset therefore receive little or no training signal, meaning that they do not evolve under gradient-based optimisation, such as the \ac{MFLD} or AdamW methods described in \Cref{sec:intro}.
It is to this issue, which we call \emph{responsibility starvation} that we attribute the difficulty of particle-based optimisation on the mixture risk, and we substantiate this theoretically in \Cref{sec:slot-theory} and empirically in \Cref{sec: experiments}.

\subsection{The Bridge}
\label{subsec: bayespro}

Our proposed remedy is simple and straightforward to implement; establishing its effectiveness is our main technical contribution in \Cref{sec:slot-theory,sec: experiments}.
The idea is to regularise the optimisation landscape in such a manner that responsibility starvation is avoided, ensuring all particles receive sufficient training signal.
To this end we construct a \emph{bridge} between the (challenging to optimise) mixture risk \eqref{eq: log PrO} and the (favourable to optimise) individual risk \eqref{eq: loss Bayes}: 

\begin{definition}[Bridge]
\label{def:bayes-pro-bridge}
For \(\beta\in[0,1]\), the \emph{bridge} risk is defined as
\begin{equation}
R_{\beta}(Q)
=
(1-\beta)R_{\mathrm{mix}}(Q)
+
\beta R_{\mathrm{ind}}(Q) .
\label{eq:distributional-bridge}
\end{equation}
\end{definition}

\noindent As a convenient shorthand we refer to $\beta=0$ as \emph{pure mixture} and to $\beta=1$ as \emph{independent training}, since in the latter case each particle can be trained in parallel.
Our focus is on values $\beta \ll 1$, for which the individual risk acts as a regulariser rather than as the dominant component.
Such regularisation is of course not novel; for example, for particle mixtures, $R_\beta(Q_m)$ is the augmented objective of \citet{jeffares2023joint}.
Similar heuristics are employed in particle swarm optimisation \citep{kennedy1995particle}, where each particle in the swarm is expected to also achieve reasonable individual performance at the optimisation task.
Rather, our contribution is to demonstrate the suitability of this particular form of regularisation for particle-based optimisation of the mixture risk.
To this end, we first explain how regularisation using the bridge prevents the vanishing responsibility issue described in \Cref{subsec:particle-mixtures}.

Let $w_{ij}=p_{\bm{\theta}_j}(y_i\mid x_i)$ denote the probability (or probability density) assigned by particle \(j\) to the observed response \(y_i\).
The negative derivative of the bridge risk with respect to the particle log-probability is
\begin{equation}
    -
    \frac{
        \partial
        R_{\beta}(Q_m)
    }{
        \partial\log w_{ij}
    }
    =
    \frac{1}{n}
    \rho_{ij}^{(\beta)},
    \qquad
    r_{ij} = \frac{
        w_{ij}
    }{
        \sum_{k=1}^{m}w_{ik}
    } , \qquad
    \rho_{ij}^{(\beta)}
    =
    (1-\beta)
    r_{ij}
    +
    \frac{\beta}{m}.
    \label{eq:convex-responsibility}
\end{equation}
To distinguish the quantities $\rho_{ij}^{(\beta)}$ from the $r_{ij}$, we will refer to the $\rho_{ij}^{(\beta)}$ as \emph{effective responsibilities}, which again capture the extent to which the $j$th particle is updated using the $i$th example in the dataset, in this case when the bridge risk is employed.
At \(\beta=0\), the effective responsibility of particle \(j\) depends on the predictions of every particle, and can vanish if the $j$th particle does not produce a meaningful prediction for the $i$th example; this is the responsibility starvation described in \Cref{subsec:particle-mixtures}.
Our main insight is that taking any $\beta > 0$ imposes a hard lower bound on each responsibility, preventing starvation, and in the extreme case where \(\beta=1\), the effective responsibility \(\rho_{ij}^{(1)}=1/m\) is uniform, so that every particle receives a similarly strong training signal.

To see why this is the case, decompose the risk function into its individual data terms as $R_\beta = \frac{1}{n}\sum_{i=1}^n \ell_{\beta,i}$, let $u_m = (\frac{1}{m},\dots,\frac{1}{m})$, and let $r_i = (r_{i1},\dots,r_{im})$.
The proof of the following result can be found in \Cref{app:proof-responsibility-barrier}:

\begin{proposition}[A Reverse-KL Responsibility Barrier]
\label{prop:responsibility-barrier}
$\ell_{\beta,i}
  =\ell_{0,i}+\beta\operatorname{KL}(u_m \Vert r_i)$.
\end{proposition}

\noindent Here $\operatorname{KL}(u_m \Vert r_i)$ is to be interpreted as the \ac{KL} divergence between the two multinomial distributions with pmfs characterised by $u_m, r_i \in [0,1]^m$.

\Cref{prop:responsibility-barrier} shows that, compared to the mixture loss $\ell_{0,i}$, the bridge loss $\ell_{\beta,i}$ includes a reverse-\ac{KL} penalty on the $r_i$ with the full-support uniform measure as its first argument.
Unlike the forward divergence $\operatorname{KL}(r_i \Vert u_m)$, whose summands $r_{ij}\log r_{ij}$ vanish as $r_{ij}\to0$, the reverse-\ac{KL} diverges whenever any single $r_{ij}\to0$.
It therefore behaves as a barrier, in the sense of interior-point optimisation, that renders a fully starved particle infinitely costly whenever $\beta>0$ while leaving balanced responsibilities unpenalised, and differentiating it produces the additive floor $\beta/m$ in $\rho_{ij}^{(\beta)}$, so that every particle receives a non-vanishing training signal. 

Implementation of particle-based optimisation based on $R_\beta$ is straightforward and no more computationally intensive compared to particle-based optimisation of the mixture risk $R_{\mathrm{mix}}$; this is because one can re-use intermediate quantities that were already calculated.
A short discussion can be found in \Cref{app:mfld-derivation}.
The remaining challenge is to establish the effectiveness of particle-based optimisation of $R_\beta$ theoretically (\Cref{sec:slot-theory}) and empirically (\Cref{sec: experiments}).

\begin{remark}[Relation to \(\alpha\)-R\'enyi]
\label{rem: reyni}
\citet{cordero2026soft} introduced a different notion of responsibilities that are aligned with their $\alpha$-R\'{e}nyi approach, of the form 
\begin{equation}
    \frac{
        w_{ij}^{\alpha}
    }{
        \sum_{k=1}^{m}
        w_{ik}^{\alpha}
    }.
    \label{eq:renyi-responsibility}
\end{equation}
Setting \(\alpha=1-\beta\) aligns the endpoints with our bridge (pure mixture at \(\alpha=1\), independent training at \(\alpha=0\)), but the interior paths differ: the \(\alpha\)-R\'enyi construction changes likelihood contrast through a power transformation, whereas the bridge combines predictive performance of the ensemble with individual performance of each particle.
\end{remark}

\section{Theoretical Assessment}
\label{sec:slot-theory}

The aim of this section is to explain why direct minimisation of the mixture risk over particles can fail (cf. \Cref{fig:mechanism-puzzle}), and why using the bridge allows this problem to be avoided.
In short, the mixture risk constrains only the aggregate predictive mixture, so there exist degenerate decompositions -- which we call \emph{slot partitions} -- in which individual particles simply memorise a fixed answer, that are almost as good in objective value as the intended solution.
As particles approach such a decomposition, the pure-mixture gradient that would correct a wrongly-decided particle vanishes, whereas the bridge supplies a non-vanishing gradient that provably escapes it in finite time.
\Cref{subsec:slot-degeneracy} presents a rigorous analysis of these phenomena in the context of a $K$-class classification task, while \Cref{subsec:warm-start} explains how warm-starting can also enable degenerate decompositions to be avoided.

\subsection{Improving the Optimisation Landscape}
\label{subsec:slot-degeneracy}

Our mathematical analysis is conducted in the setting of a $K$-class classification task, and we introduce the set-up and notation in \Cref{subsec: notation}.
The degenerate decompositions -- which we call \emph{slot partitions} -- which lead to problems for optimisation of the mixture risk are defined in \Cref{subsec: slot}.
Our main mathematical results, which concern the behaviour of the mixture risk and the bridge risk in the neighbourhood of slot partitions, are presented and discussed in \Cref{subsec: approaching}.

\subsubsection{Set-Up and Notation}
\label{subsec: notation}

Our analysis is performed for $K$-class classification, with a model of softmax form
\begin{equation}
    p_{\bm{\theta}}(c \mid x)
    =
    \frac{
        \exp\{s_{\bm{\theta}}(c \mid x)\}
    }{
        \sum_{c' \in\mathcal C}
        \exp\{s_{\bm{\theta}}(c' \mid x)\}
    } ,
    \qquad c \in \mathcal{C},
    \label{eq:softmax}
\end{equation}
for a parametric score function $s_{\bm{\theta}}( \cdot \mid x) : \mathcal{C} \to \mathbb{R}$.
Since the precise parametrisation can vary greatly from model-to-model, we in fact analyse the particles at the level of their score functions $s_j(\cdot | x) = s_{\bm{\theta}_j}(\cdot | x)$, treating these as \emph{unrestricted functions} of the input $x$, as would be the case for a sufficiently flexible model.
Similarly we let $p_j(\cdot\mid x)=p_{\bm{\theta}_j}(\cdot\mid x)$ denote the probabilistic classification model.

To quantify the training signal received by each particle, we let $\bm{s}_{ij}=(s_j(c\mid x_i))_{c\in\mathcal{C}}\in\mathbb{R}^K$ and $\bm{p}_{ij}=(p_j(c\mid x_i))_{c\in\mathcal{C}}\in\mathbb{R}^K$ denote the vectors of scores and probabilities output by particle $j$ for example $i$, and let $\bm{e}_c\in\mathbb{R}^K$ denote the indicator vector of the label $c\in\mathcal{C}$.
Recalling that $w_{ij}=p_j(y_i\mid x_i)$, the $i$th data term in the bridge risk \eqref{eq:distributional-bridge} is
\[
\ell_{\beta,i}=-(1-\beta)\log\left(\frac{1}{m}\sum_{k=1}^m w_{ik}\right)-\frac{\beta}{m}\sum_{k=1}^m\log w_{ik} ,
\]
and since $\nabla_{\bm{s}_{ij}}\log w_{ij}=\bm{e}_{y_i}-\bm{p}_{ij}$ from \eqref{eq:softmax}, we obtain
\begin{equation}
    \nabla_{\bm{s}_{ij}}\ell_{\beta,i}
    =\rho_{ij}^{(\beta)}\left(\bm{p}_{ij}-\bm{e}_{y_i}\right) ,
    \label{eq:score-gradient}
\end{equation}
where $\rho_{ij}^{(\beta)}$ is the effective responsibility defined in \eqref{eq:convex-responsibility}.
That is, the size of the training signal that particle $j$ receives from example $i$ is proportional to its responsibility for that example.

\subsubsection{Slot Partitions}
\label{subsec: slot}

The degenerate decompositions to which we attribute the failure of particle-based optimisation of the mixture risk will be called \emph{slot partitions}:

\begin{definition}[Slot Partition]
\label{def: slot partition}
A \emph{slot partition} assigns to each particle $j$ and each input $x$ a single label $c_j(x)\in\mathcal{C}$, its \emph{slot}, which that particle predicts with certainty; i.e. $p_j(c\mid x)=\mathbf{1}\{c=c_j(x)\}$.    
\end{definition}

\noindent The main intuition that underpins the following analysis is that, at $\beta=0$, the mixture risk \eqref{eq: discretise Rmix} can be (approximately) minimised by any of the many slot partitions that lead to (approximately) the same optimal mixture model, so that the optimisation landscape is likely to be extremely complicated.

\subsubsection{Approaching a Slot Partition}
\label{subsec: approaching}

To study what happens as the particles \emph{approach} a slot partition, we work with the \emph{slot-margin family}
\begin{equation}
    s_j(c\mid x)=M\,\mathbf{1}\{c=c_j(x)\}
    \qquad\text{for all } x \text{ and all } c\in\mathcal{C},
    \label{eq:slot-margin-main}
\end{equation}
so that, from \eqref{eq:softmax}, particle $j$ assigns probability $e^M/(e^M+K-1)$ to its slot label $c_j(x)$, and probability $1/(e^M+K-1)$ to each of the other $K-1$ labels.
Thus $M\to0$ corresponds to uniform predictions, while $M\to\infty$ recovers the slot partition in the limit.
We let $m_c(x)=|\{j : c_j(x)=c\}|$ denote the number of particles whose slot at input $x$ is the label $c$.

The proof of the following result is contained in \Cref{app: proof gradients}:

\begin{proposition}[Escaping from Slot Partitions]
\label{thm:slot-degeneracy}

Suppose that, for every example $i$, at least one particle's slot is the correct label ($m_{y_i}(x_i)\geq1$), and let $m_{\min}=\min_{i}m_{y_i}(x_i)$. 

\begin{enumerate}
\item \emph{(Pure mixture has vanishing gradient)}
Along the slot-margin family \eqref{eq:slot-margin-main}, the pure-mixture gradient satisfies
\begin{equation}
  \max_{i,j}\left\|\nabla_{\bm{s}_{ij}}\ell_{0,i}\right\|_1
  \leq\frac{2(K-1)}{m_{\min}e^M}
  \longrightarrow0
  \qquad(M\to\infty),
  \label{eq:slot-pure-gradient-bound}
\end{equation}
so pure mixture exerts a vanishing corrective force as the particles approach a slot partition, however wrong each individual particle is. 
\item \emph{(The bridge has non-vanishing gradient)}
For every fixed $\beta>0$, and every example $i$ and particle $j$ whose slot is wrong for that example ($c_j(x_i)\neq y_i$), the gradient instead converges to a strictly positive limit,
\begin{equation}
  \left\|\nabla_{\bm{s}_{ij}}\ell_{\beta,i}\right\|_1
  \longrightarrow\frac{2\beta}{m}>0
  \qquad(M\to\infty).
  \label{eq:slot-beta-boundary-gradient}
\end{equation}
Moreover, for $\beta>0$, consider the gradient flow $\dot{\bm{s}}_{ij}=-\nabla_{\bm{s}_{ij}}\ell_{\beta,i}$ for a single example $i$ and a particle $j$ with $c_j(x_i)\neq y_i$, started from a wrong-slot margin $M_0=s_j(c_j(x_i)\mid x_i)-s_j(y_i\mid x_i)>0$. Fix $\varepsilon\in(0,\tfrac12)$. Then, for as long as particle $j$ assigns probability at least $1-\varepsilon$ to $c_j(x_i)$ and at most $\varepsilon$ to $y_i$, the margin decreases at rate at least $2\beta(1-\varepsilon)/m$; in particular, particle $j$ leaves this neighbourhood of its slot within time
$T_{\mathrm{esc}}\leq mM_0/(2\beta(1-\varepsilon))$.
\end{enumerate}
\end{proposition}

\noindent This is the mechanism behind the failure of direct particle-based optimisation of the mixture risk observed in earlier work: under the mixture risk, a decomposition in which particles collapse onto individual answer slots is almost as effective as the intended balanced solution, and once particles approach it, no gradient corrects it (part 1 of \Cref{thm:slot-degeneracy}).
Any positive $\beta$ removes this obstruction, by supplying a gradient that escapes a degenerate slot partition in finite time (part 2 of \Cref{thm:slot-degeneracy}).

\begin{remark}[Beyond Classification]
The slot-margin family is specific to the classification setting, but the underlying mechanism is not.
For an arbitrary model $p_{\bm{\theta}}(y\mid x)$, it follows from \eqref{eq:convex-responsibility} that
\[
\nabla_{\bm{\theta}_j} R_\beta(Q_m) = -\frac{1}{n}\sum_{i=1}^n \rho_{ij}^{(\beta)} \nabla_{\bm{\theta}} \log p_{\bm{\theta}_j}(y_i\mid x_i),
\]
so that at $\beta=0$ example $i$ contributes to the update of particle $j$ only in proportion to the responsibility $r_{ij}$.
For instance, in regression with Gaussian predictive densities $p_{\bm{\theta}_j}(y\mid x)=\mathcal{N}(y;\mu_j(x),\sigma^2)$, one has
\[
r_{ij}\leq\frac{w_{ij}}{w_{ik}}=\exp\left\{-\frac{(y_i-\mu_j(x_i))^2-(y_i-\mu_k(x_i))^2}{2\sigma^2}\right\}
\]
for every particle $k$, which is exponentially small whenever particle $j$ predicts $y_i$ poorly but some other particle $k$ predicts it well.
\Cref{prop:responsibility-barrier} holds in this setting, so that for any $\beta>0$ the floor $\rho_{ij}^{(\beta)}\geq\beta/m$ is retained.
\end{remark}

\begin{remark}[Limitations of the Analysis]
There are two main limitations to the analysis which we have just presented:
\begin{enumerate}
\item Our analysis concerned the training signal on the scores of each example, assuming a sufficiently flexible parametric model.
In practice, the scores for different examples are coupled through the shared parameters $\bm{\theta}$.
As such, \Cref{thm:slot-degeneracy} is not a global convergence statement for the parameters themselves, rather it serves as a phenomenological explanation for the empirical behaviour reported in earlier work and reaffirmed in \Cref{sec:case-pro,sec:experiments}.
\item Part 2 of \Cref{thm:slot-degeneracy} concerns gradient flow; adaptive optimisers such as AdamW rescale a uniformly small gradient to a step of fixed size, so that the magnitude bound in part 1 need not translate into a slow escape when AdamW is used.
However, such rescaling does not alter the \emph{relative} size of the effective responsibilities $\rho_{ij}^{(\beta)}$, and it is this relative weighting that the bridge is designed to control.
\end{enumerate}
\end{remark}

\subsection{All You Need is a Competent Initialisation}
\label{subsec:warm-start}

One might wonder whether a \emph{warm start}, i.e. initialisation sufficiently close to an optimal solution, would also resolve the problem; the answer turns out to be positive provided that the individual models are competent.
The proof of the following result is contained in \Cref{app:proof-competence-floor}.

\begin{proposition}[Competence-Induced Responsibility Floor]
\label{cor:competence-floor}
On every pair with $w_{ij}\geq\kappa$,
\begin{equation}
  r_{ij}\geq\frac{\kappa}{m},
  \qquad
  \rho_{ij}^{(\beta)}
  \geq\frac{\beta+(1-\beta)\kappa}{m}.
  \label{eq:slot-implicit-floor}
\end{equation}
Thus beginning optimisation from a model with larger correct-label probabilities $w_{ij}$ imposes an implicit responsibility floor, even at $\beta=0$.
\end{proposition}
\noindent The proof of \Cref{cor:competence-floor} uses only the fact that $w_{ij}\leq1$, which holds whenever the response is discrete; for a continuous response whose predictive densities are bounded above by $B$, the same argument gives $r_{ij}\geq\kappa/(mB)$.

\Cref{cor:competence-floor} also indicates how the choice of $\beta$ should depend on the initialisation: the floor in \eqref{eq:slot-implicit-floor} is the sum of an explicit contribution $\beta/m$ and an implicit contribution $(1-\beta)\kappa/m$, so that the more competent the initial model (i.e. the larger $\kappa$), the smaller the value of $\beta$ needed to guarantee a given responsibility floor.
In Case Study~2 a warm start corresponds to a more competent \ac{LLM} backbone, and the implications of \Cref{cor:competence-floor} are tested in \Cref{subsec:exp-competence}.
More generally, any warm start whose particles already assign non-negligible probability to the observed responses, such as initialisation from a pretrained model or from a preliminary fit, confers an implicit responsibility floor even on the pure mixture risk, and correspondingly reduces the weight on the individual risk that is required.

\section{Empirical Assessment}
\label{sec: experiments}

Our empirical assessment aims both to demonstrate the failure of particle-based optimisation of the mixture risk, and to understand the extent to which this failure can be avoided using our proposed method.
This is achieved across two case studies, aligned to our two motivating examples in \Cref{sec:intro}:
\Cref{sec:case-pro} concerns computation of a \ac{PrO} posterior using \ac{MFLD}, while \Cref{sec:experiments} concerns the joint fine-tuning of \acp{LLM} using noisy AdamW.

\subsection{Case Study 1: Computing PrO Posteriors}
\label{sec:case-pro}

This case study considers computation of the \ac{PrO} posterior $Q_{\mathrm{PrO}}$, the minimiser of \eqref{eq: obj} under the mixture risk $R_{\mathrm{mix}}$.
First we characterise the bridge at the distributional level in this \ac{KL}-regularised context (\Cref{subsec:cs1-bridge}), before conducting an empirical assessment (\Cref{subsec:cs1-numerics}), and finally we explain how the solution depends on $\beta$ (\Cref{subsec:cs1-phase}).

\subsubsection{The Bridge at the Distributional Level}
\label{subsec:cs1-bridge}

An application of the bridge to \eqref{eq: obj} corresponds to the regularised objective
\begin{equation}
J_{\beta}(Q)
=
R_{\beta}(Q)
+
\frac{1}{n}\operatorname{KL}(Q\Vert\Pi) ,
\label{eq:bridge-objective}
\end{equation}
and we let \smash{$Q_\beta^{(n)}$} denote its minimiser.
Thus \smash{$Q_0^{(n)}$} is the \ac{PrO} posterior, while \smash{$Q_1^{(n)}$} is the standard Bayesian posterior.
It turns out that the intermediate distributions $Q_\beta^{(n)}$ also admit a statistical interpretation, which we explain in terms of the \(\beta\)-tempered Bayesian reference measure $\Pi_{n,\beta}$ defined via the Radon--Nikodym derivative
\begin{equation}
\frac{\mathrm d\Pi_{n,\beta}}
     {\mathrm d\Pi}(\bm{\theta})
=
\frac{1}{Z_{n,\beta}}
\prod_{i=1}^{n}
p_{\bm{\theta}}(y_i\mid x_i)^{\beta},
\qquad
Z_{n,\beta}
=
\int
\prod_{i=1}^{n}
p_{\bm{\theta}}(y_i\mid x_i)^{\beta}
\,\mathrm d\Pi(\bm{\theta}),
\label{eq:tempered-reference-measure}
\end{equation}
whenever \(0<Z_{n,\beta}<\infty\).
The proof for the following result is contained in \Cref{app:proof-tempered-reference}:

\begin{proposition}[Tempered Representation]
\label{prop:tempered-reference}

Assume \(0<Z_{n,\beta}<\infty\). 
Then
\begin{equation}
Q_\beta^{(n)}
=
\argmin_{Q \in \mathcal{P}(\mathbb{R}^d)}
\left\{
(1-\beta)
R_{\mathrm{mix}}(Q)
+
\frac{1}{n}
\operatorname{KL}
\left(
Q\Vert\Pi_{n,\beta}
\right)
\right\}.
\label{eq:tempered-reference-argmin}
\end{equation}
\end{proposition}

\noindent Thus for \(\beta\in(0,1)\), use of the bridge can be interpreted as regularising \(Q\) towards a tempered Bayesian posterior $\Pi_{n,\beta}$ instead of the original reference measure $\Pi$.

\begin{remark}[Comparison to Predictive Variational Inference]
\citet{lai2024predictive} proposed 
\begin{equation}
\argmin_{Q \in \mathcal{P}(\mathbb{R}^d)}
\left\{
R_{\mathrm{mix}}(Q)
+
\frac{1}{n}
\operatorname{KL}
\left(
Q\Vert\Pi_{n,1}
\right)
\right\} , \label{eq: lai}
\end{equation}
which can be interpreted as the \ac{PrO} posterior in which the standard Bayesian posterior $\Pi_{n,1}$ acts as the reference distribution.
Since $\operatorname{KL}(Q\Vert\Pi_{n,1})=\operatorname{KL}(Q\Vert\Pi)+nR_{\mathrm{ind}}(Q)+\log Z_{n,1}$, the objective in \eqref{eq: lai} is equal, up to an additive constant, to $2 R_{1/2}(Q)+\frac{1}{n}\operatorname{KL}(Q\Vert\Pi)$.
Thus \eqref{eq: lai} implicitly uses the bridge with $\beta = 1/2$ fixed, and assigns twice as much importance to the data term compared to the standard \ac{PrO} posterior $Q_{\mathrm{PrO}}$.
\end{remark}

\subsubsection{Numerical Results}
\label{subsec:cs1-numerics}

Now we report a synthetic experimental assessment of the effectiveness of the bridge in the context of computing a \ac{PrO} posterior:

\paragraph{Task}

We use the data-generating process of \citet[Section 5.2]{mclatchie2025predictively}, who found the \ac{PrO} posterior to differ materially from the Bayesian posterior.
The covariates $x=(x^{(0)},x^{(1)})$ are uniform on $[-2,2]^2$, the label is $y=0$ when $x^{(0)}<0$ and $x^{(1)}>0$, $y=1$ when $x^{(0)}>0$ and $x^{(1)}<0$, and $y\sim\mathrm{Bernoulli}(0.5)$ in the two remaining quadrants.
The model is logistic regression with an intercept, $p_{\bm{\theta}}(y=1\mid x)=\sigma(\theta_0+\theta_1x^{(0)}+\theta_2x^{(1)})$, so that $d=3$, with prior $\Pi=\mathcal N(0,100\,\bm{I})$, $n=1{,}000$ training examples, $10{,}000$ held-out examples, and the weight $n$ on the risk in \eqref{eq: obj}.
No single logistic regression fits both the deterministic and the random quadrants, whereas an equally weighted mixture of a steep logistic regression along $+x^{(0)}$ and one along $-x^{(1)}$ recovers the data-generating distribution, denoted $P_0$, in the limit of large slopes.
The task therefore has both features that \Cref{sec:slot-theory} identifies as conducive to starvation: each particle shares its parameters across all examples, and the mixture predictive can be decomposed across particles in many ways.

\paragraph{Methods Compared}
We run the \ac{MFLD} of \eqref{eq:intro-mfld} using the bridge $R_\beta$ in place of the mixture risk $R_{\mathrm{mix}}$.
For this experiment we used $m=32$ particles, Euler--Maruyama step size $\epsilon=10^{-4}$ and $2\times10^{6}$ iterations, that is up to \ac{MFLD} time $t=200$.
Eight independent realisations were simulated to account for variability due to the random seed.

The particles are initialised either by independent draws from $\Pi$, or at a slot partition (\Cref{subsec:slot-degeneracy}) in which half of the particles have intercept $+M$ and half $-M$, with zero slopes and $M\in\{8,16\}$, so that every particle predicts one label with confidence on every input.
Further runs with $m\in\{8,128\}$, $n\in\{200,5{,}000\}$ and chains of $10^{7}$ iterations are described in \Cref{app:mfld-implementation}.

\paragraph{Reference Solution}

For every $\beta>0$ the particle system converges: the chains agree from all three initialisations, with a standard deviation across chains of the held-out \ac{NLL} of at most $0.0005$, and runs with $m=128$ reproduce the held-out \ac{NLL} of $m=32$ to within $0.0002$.
For $\beta=0$ the chains reach no common solution, also with $m=128$ and after $10^{7}$ iterations, and this lack of agreement across chains and initialisations is the quantity we report.

\paragraph{Diagnostics}

We report the training-set value $R_{\mathrm{mix}}(Q^t_m)$, the smallest raw responsibility $r_{ij}$ over all examples and particles, and the held-out \ac{NLL} of the mixture predictive.
We also report the \ac{MFLD} time until the gap between $R_{\mathrm{mix}}(Q^t_m)$ and its lowest population value (i.e. the conditional entropy $H(Y\mid X)=\tfrac12\log2$) has halved, and, for $m=8$, whether the particles have reached the decomposition that recovers the true data-generating distribution, with four particles steep along $+x^{(0)}$ and four along $-x^{(1)}$.

\begin{figure}[!tbp]
\centering
\includegraphics[width=\textwidth]{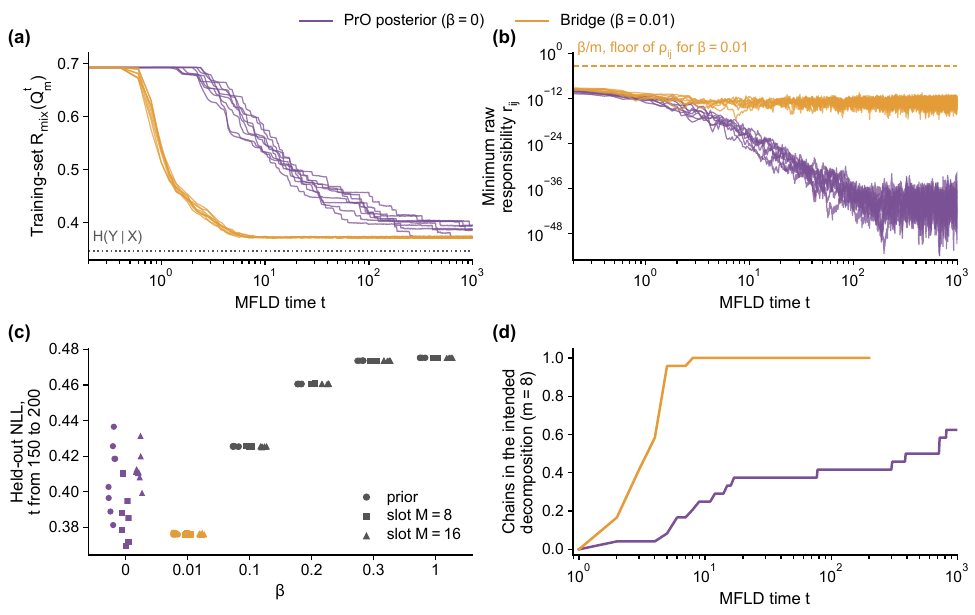}
\caption{Case Study~1 on the misspecified logistic regression of \citet{mclatchie2025predictively}, with $n=1{,}000$. The \ac{PrO} posterior ($\beta=0$) is in purple and the bridge with $\beta=0.01$ in orange. (a) Training-set $R_{\mathrm{mix}}$ of eight chains with $m=32$, started from a slot partition with $M=16$ and run to $t=1{,}000$, with the lowest population value $H(Y\mid X)$ dotted. (b) Minimum raw responsibility $\min_{i,j}r_{ij}$ of the same chains. The effective responsibility of the bridge is at least $\beta/m$ (dashed) by construction. (c) Held-out \ac{NLL} of every chain with $m=32$, averaged over $t$ from $150$ to $200$, for the six values of $\beta$ and the three initialisations, with $\beta\geq0.1$ in grey. (d) Fraction of the $24$ chains with $m=8$, eight per initialisation, in which four particles are steep along $+x^{(0)}$ and four along $-x^{(1)}$, the decomposition that recovers $P_0$. The chains at $\beta=0$ are continued to $t=1{,}000$, and those at $\beta=0.01$ are shown to $t=200$.}
\label{fig:cs1}
\end{figure}

\paragraph{Findings}

At $\beta=0$, \ac{MFLD} fails to converge at the level of the held-out \ac{NLL} (\Cref{fig:cs1}c).
Indeed, the held-out \ac{NLL} depends on the initialisation, $0.4086\pm0.0192$ from the prior, $0.3838\pm0.0140$ from the slot partition with $M=8$ and $0.4130\pm0.0093$ with $M=16$ (\Cref{tab:cs1} in \Cref{app: additional 1}).
In contrast, for every $\beta>0$ the three initialisations agree to within $0.0005$.
From the slot partition with $M=16$, the gap in $R_{\mathrm{mix}}$ halves after \ac{MFLD} time $13.5$ at $\beta=0$, against $1.08$ at $\beta=0.01$ (\Cref{fig:cs1}a) and $0.15$ at $\beta=0.1$ (\Cref{tab:cs1}).
Neither more particles nor longer chains repair it:
With $m=128$ particles, the three initialisations at $\beta=0$ give $0.4137$, $0.3835$ and $0.4137$, and after $10^{7}$ iterations the chains with $m=32$ still range from $0.369$ to $0.427$, against $0.376$ to $0.377$ at $\beta=0.01$.

The mechanism behind this failure at $\beta = 0$ is indeed the degenerate decompositions discussed in \Cref{sec:slot-theory}.
Raw responsibilities fall to about $10^{-41}$ at $\beta=0$, whereas the bridge keeps every effective responsibility at least $\beta/m$ (\Cref{prop:responsibility-barrier} and \Cref{fig:cs1}b).
With $m=8$, chains at $\beta=0$ settle in decompositions in which some particles are steep in neither of the two directions that are characteristic of the true data-generating process.
These $34$ of the $192$ particles misclassify $45\%$ of the training examples, against the $25\%$ that any steep particle misclassifies in the random quadrants.
On $94\%$ of the examples they misclassify, their raw responsibility is below $0.01$, so that the signal that would reorient them is $40$ times below the uniform share $1/m$; the optimisation path is stuck in the neighbourhood of a slot partition as described in \Cref{thm:slot-degeneracy}.
At $\beta=0$, $14$ of the $24$ chains are trapped in such a decomposition at $t=200$ and $9$ remain trapped at $t=1{,}000$, whereas at $\beta=0.01$ all $24$ chains reach the decomposition that recovers the true data-generating process by $t=10$ (\Cref{fig:cs1}d).

A small $\beta$ retains the benefit of the \ac{PrO} posterior:
The small number of chains at $\beta=0$ with $m=8$ that reach the intended decomposition attain a held-out \ac{NLL} of $0.362$, against $0.376$ from \emph{every} initialisation of the bridge at $\beta = 0.01$.
It is also the best calibrated setting, with an expected calibration error (ECE) of $0.030$, against $0.038$ to $0.063$ at $\beta=0$ and between $0.045$ and $0.054$ for $\beta\geq0.1$ (\Cref{tab:cs1} in \Cref{app: additional 1}).
Larger values of $\beta$ move $Q^{(n)}_\beta$ towards the tempered Bayesian posterior of \Cref{prop:tempered-reference}, with held-out \ac{NLL} between $0.425$ and $0.475$, so that in this task we advocate for use of the bridge with $\beta=0.01$, in line with reporting $Q^{(n)}_\beta$ with a small fixed $\beta$ as a regularised \ac{PrO} posterior (\Cref{subsec:cs1-bridge}).

\subsubsection{A Phase Transition in \texorpdfstring{$\beta$}{beta}}
\label{subsec:cs1-phase}

To understand how we can choose $\beta$, and still obtain the predictive benefits of the \ac{PrO} posterior, we present an informal local analysis which predicts a phase transition at a critical value $\beta_{\mathrm{crit}}$. This analysis demonstrates that using the bridge at values of $\beta> \beta_{\mathrm{crit}}$ can produce worse predictive results than the original \ac{PrO} posterior.

First note that, since the weight $\frac{1}{n}$ on the \ac{KL} term in \eqref{eq:bridge-objective} vanishes as $n \rightarrow \infty$, the large-sample behaviour of \smash{$Q_\beta^{(n)}$} is governed by the population risks $\mathcal{R}_\beta$, $\mathcal{R}_{\mathrm{mix}}$ and $\mathcal{R}_{\mathrm{ind}}$, defined as $R_\beta$, $R_{\mathrm{mix}}$ and $R_{\mathrm{ind}}$ but with the average over the dataset replaced by an expectation $\mathbb{E}$ over $(X,Y) \sim P_0$.
Let $\bm{\theta}^\star$ minimise $\bm{\theta} \mapsto \mathcal{R}_{\mathrm{ind}}(\delta_{\bm{\theta}})$, so that $\delta_{\bm{\theta}^\star}$ is the large-sample limit of the Bayesian posterior in regular settings (cf. \Cref{sec:intro}).

\paragraph{Identification of $\beta_{\mathrm{crit}}$}

Fix $(x,y)$ and, for small $\tau>0$ and direction $\bm{v}\in\mathbb{R}^d$, consider splitting the point mass $\delta_{\bm{\theta}^\star}$ across two particles: $Q_\tau = \frac{1}{2} \delta_{\bm{\theta}^\star + \tau \bm{v}} + \frac{1}{2} \delta_{\bm{\theta}^\star - \tau \bm{v}}$. 
A second-order Taylor expansion of  $\log p_{\bm{\theta}^\star \pm \tau \bm{v}}(y|x)$ and $p_{\bm{\theta}^\star \pm \tau \bm{v}}(y|x)$ in $\tau$ yields 
\begin{align*}
\int \log p_{\bm{\theta}}(y|x) \; \mathrm{d}Q_\tau(\bm{\theta}) & = \log p_{\bm{\theta}^\star}(y|x) + \frac{\tau^2}{2} b + o(\tau^2), \\
\log \int p_{\bm{\theta}}(y|x) \; \mathrm{d}Q_\tau(\bm{\theta}) & = \log p_{\bm{\theta}^\star}(y|x) + \frac{\tau^2}{2} \left( b + a^2 \right) + o(\tau^2) .
\end{align*}where the first-order terms in both expansions cancel when averaging over particles, and where
$$
a = \bm{v}^\top \nabla_{\bm{\theta}} \log p_{\bm{\theta}^\star}(y|x) \; \text{ and } \; b = \bm{v}^\top \nabla_{\bm{\theta}}^2 \log p_{\bm{\theta}^\star}(y|x) \bm{v}.
$$
Both risks therefore incur the same curvature penalty, $-b$, but the mixture risk used by \ac{PrO} additionally \emph{rewards} spreading by $a^2$, since an example with $a \neq 0$ is better predicted by one of the two particles.
Taking expectations, and letting $\bm{H} = - \mathbb{E}[ \nabla_{\bm{\theta}}^2 \log p_{\bm{\theta}^\star}(Y|X) ]$ and $\bm{G} = \mathbb{E}[ \nabla_{\bm{\theta}} \log p_{\bm{\theta}^\star}(Y|X) \nabla_{\bm{\theta}} \log p_{\bm{\theta}^\star}(Y|X)^\top ]$, we obtain
\begin{align}
\mathcal{R}_\beta(Q_\tau) = \mathcal{R}_\beta(\delta_{\bm{\theta}^\star}) + \frac{\tau^2}{2} \bm{v}^\top \left( \bm{H} - (1-\beta) \bm{G} \right) \bm{v} + o(\tau^2) . \label{eq:cs1-local}
\end{align}
Thus, assuming $\bm{H}$ and $\bm{G}$ are positive definite, the point mass $\delta_{\bm{\theta}^\star}$ can be improved by a small spread, to second order, if and only if $\beta < \beta_{\mathrm{crit}}$, where
\begin{align}
\beta_{\mathrm{crit}} = \max\left\{ 0 , 1 - \frac{1}{\gamma} \right\} , \qquad \gamma = \max_{\bm{v} \neq \bm{0}} \frac{\bm{v}^\top \bm{G} \bm{v}}{\bm{v}^\top \bm{H} \bm{v}} . \label{eq:beta-crit}
\end{align}
For $\beta < \beta_{\mathrm{crit}}$ we therefore expect \smash{$Q_\beta^{(n)}$} to remain spread out as $n \rightarrow \infty$, as does the \ac{PrO} posterior, while for $\beta > \beta_{\mathrm{crit}}$ we expect it to concentrate around $\bm{\theta}^\star$, as does the Bayesian posterior (though, since the analysis is local, $\delta_{\bm{\theta}^\star}$ need not then minimise $\mathcal{R}_\beta$).
A similar transition is exhibited by the $\alpha$-R\'{e}nyi risk of \citet{cordero2026soft}.

\paragraph{Empirical Confirmation}

\begin{figure}[!tbp]
\centering
\includegraphics[width=\textwidth]{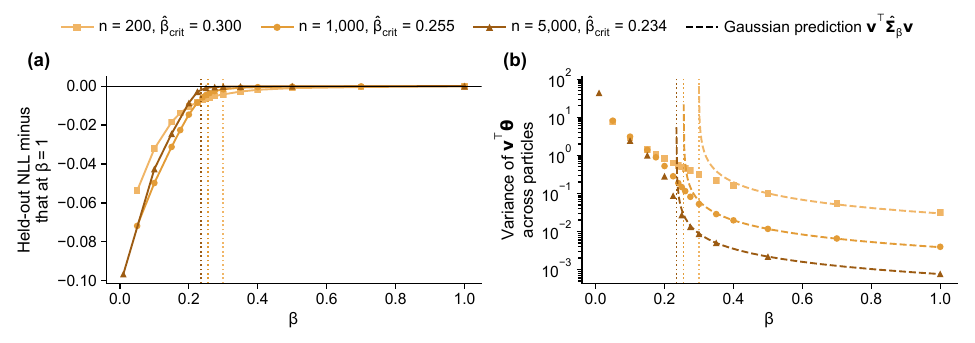}
\caption{Phase transition in $\beta$ for Case Study~1, with $n\in\{200,1{,}000,5{,}000\}$. The \ac{MFLD} for the bridge was run with $m=32$ particles and eight chains initialised from $\Pi$, to time $t=40$ for $n\in\{200,1{,}000\}$ and $t\geq10$ for $n=5{,}000$, and every point is an average over the second half of each chain. (a) Held-out \ac{NLL} minus that of the Bayesian posterior ($\beta=1$) at the same $n$. (b) Variance of $\bm{v}^\top\bm{\theta}$ across the particles, where the unit vector $\bm{v}$ attains the maximum in \eqref{eq:beta-crit}, together with the Gaussian prediction $\bm{v}^\top\hat{\bm{\Sigma}}_\beta\bm{v}$ (dashed) for $\beta>\hat{\beta}_{\mathrm{crit}}$, where $\hat{\bm{\Sigma}}_\beta$ is \eqref{eq:cs1-sigma} with $\hat{\bm{H}}$, $\hat{\bm{G}}$ and $\hat{\bm{\theta}}$ in place of $\bm{H}$, $\bm{G}$ and $\bm{\theta}^\star$. Dotted lines mark $\hat{\beta}_{\mathrm{crit}}$. }
\label{fig:cs1-phase}
\end{figure}

\Cref{fig:cs1-phase} confirms this prediction.
We ran the \ac{MFLD} of \Cref{subsec:cs1-numerics} on a grid of values of $\beta$ for $n \in \{200, 1{,}000, 5{,}000\}$, and computed $\hat{\bm{H}}$, $\hat{\bm{G}}$ and $\hat{\beta}_{\mathrm{crit}}$ from the $n$ training examples, with the mode $\hat{\bm{\theta}}$ of the Bayesian posterior in place of $\bm{\theta}^\star$.
This gives $\hat{\beta}_{\mathrm{crit}} = 0.300$, $0.255$ and $0.234$ for the three values of $n$, and $0.227$ with $200{,}000$ examples.
For $\beta > \beta_{\mathrm{crit}}$ the matrix $\bm{H} - (1-\beta)\bm{G}$ is positive definite, and repeating the expansion that led to \eqref{eq:cs1-local} for $Q = \mathcal{N}(\bm{\theta}^\star, \bm{\Sigma})$ in \eqref{eq:bridge-objective}, then minimising over $\bm{\Sigma}$, gives the covariance
\begin{align}
\bm{\Sigma}_\beta = \left[ n \left( \bm{H} - (1-\beta) \bm{G} \right) - \nabla^2 \log \pi(\bm{\theta}^\star) \right]^{-1} , \label{eq:cs1-sigma}
\end{align}
which is of order $1/n$, and which at $\beta = 1$ is the covariance of the Laplace approximation of the Bayesian posterior.
Both regimes appear as predicted.
Above $\hat{\beta}_{\mathrm{crit}}$ the particles concentrate, and their variance along the direction $\bm{v}$ that attains $\gamma$ follows $\bm{v}^\top \hat{\bm{\Sigma}}_\beta \bm{v}$.
At $\beta = 0.5$, for every $n$, the two agree to within $1\%$ and the held-out \ac{NLL} is within $0.002$ of that of the Bayesian posterior.
Below $\hat{\beta}_{\mathrm{crit}}$ the particles remain spread out, with variance $2.8$, $3.1$ and $2.4$ at $\beta = 0.1$ for the three values of $n$, and the bridge improves on the held-out \ac{NLL} of the Bayesian posterior by a margin that grows as $\beta$ decreases, reaching $0.097$ at $\beta = 0.01$ for $n = 5{,}000$.
The transition is continuous, and it sharpens as $n$ grows.
The value $\beta = 0.01$ used above therefore lies deep inside the range that retains the benefit of the \ac{PrO} posterior, and the threshold explains why $\beta = 0.3$ matches the Bayesian posterior in \Cref{fig:cs1}(c).

\subsection{Case Study 2: Joint Training of LLM Ensembles}
\label{sec:experiments}

This case study concerns an ensemble of $m$ \acp{LoRA}, attached to a shared frozen \ac{LLM} backbone, trained jointly for multiple-choice reasoning by minimising the bridge risk $R_\beta(Q_m)$.
The theoretical analysis of \Cref{sec:slot-theory} makes four predictions which we seek to test:
\begin{enumerate}
\item[(H1)] \emph{Pathology of the mixture risk.} Pure mixture training attains a worse value of its own risk $R_{\mathrm{mix}}$ than the bridge with a small $\beta>0$, on both training and held-out data and across backbones.
\item[(H2)] \emph{The mechanism is responsibility starvation.} Under pure mixture training, the particles drift towards slot-partition-like configurations in which some raw responsibilities $r_{ij}$ become vanishingly small (\Cref{thm:slot-degeneracy}).
\item[(H3)] \emph{Positive $\beta$ escapes starvation.} Switching on $\beta>0$ from a starved state restores a non-vanishing training signal (part 2 of \Cref{thm:slot-degeneracy}).
\item[(H4)] \emph{Backbone competence matters.} The benefit of $\beta>0$, and the size of $\beta$ that is required, decrease as the backbone becomes more competent (\Cref{cor:competence-floor}).
\end{enumerate}
After describing the model and experimental protocol in \Cref{subsec:protocol}, we address (H1) in \Cref{subsec:exp-puzzle}, (H2) and (H3) in \Cref{subsec:exp-mechanism}, and (H4) in \Cref{subsec:exp-competence}, before comparing against the $\alpha$-R\'{e}nyi approach of \citet{cordero2026soft} in \Cref{subsec:exp-renyi}.

\subsubsection{Model and Protocol}
\label{subsec:protocol}

Here we explain the protocol for our detailed experimental assessment of the effectiveness of the bridge in the context of jointly fine-tuning \acp{LLM}:

\paragraph{Task}
We consider fine-tuning of \acp{LLM} to solve multiple choice problems presented in plain text, where the input $x_i$ is the question and $y_i$ labels the correct element of the candidate answer set.
Since updating all parameters is prohibitive when $m$ particles must be trained and stored simultaneously, we use parameter-efficient fine-tuning \citep{hu-etal-2023-llm} with \acp{LoRA} \citep{hu2022lora}, which freeze each targeted weight matrix $\bm{W}_0$ and learn a low-rank update $\bm{W}_0 \mapsto \bm{W}_0 + \bm{B}\bm{A}$, reducing the number of trainable parameters by orders of magnitude.
Each particle $\bm{\theta}$ is a \ac{LoRA}, and the output of the fine-tuned \ac{LLM} for example $i$ is a probability distribution $p_{\bm{\theta}}(c | x_i)$ over the candidate label set $c \in \mathcal{C}_i$.
To construct it, the input \(x_i\) is written into a text prompt \(h_i\) that lists the candidates, each named by a single letter token \(t_c\) (A, B, \dots), and asks for the letter of the correct answer.
Let \(\mathrm{LM}_{\bm{\theta}}(t \mid h)\) denote the probability that the language model, with \ac{LoRA} \(\bm{\theta}\), assigns to \(t\) as the next token after the text \(h\).
The class-label probability is then the softmax of the resulting scores over the candidate set,
\begin{equation}
    p_{\bm{\theta}}(c \mid x_i)
    =
    \frac{
        \exp\!\left\{s_{\bm{\theta}}(c \mid x_i)\right\}
    }{
        \sum_{c' \in\mathcal C_i}
        \exp\!\left\{s_{\bm{\theta}}(c' \mid x_i)\right\}
    },
    \qquad
    s_{\bm{\theta}}(c \mid x_i)
    =
    \log
    \mathrm{LM}_{\bm\theta}
    \bigl(
        t_c \mid h_i
    \bigr).
    \label{eq:class-probability}
\end{equation}
This is a softmax classifier of the form \eqref{eq:softmax} analysed in \Cref{sec:slot-theory}, with scores given by the log next-token probabilities of the letter tokens.

\paragraph{Backbones}
We consider four backbones: Llama 3.2-1B and Llama 3.2-3B~\citep{meta2024llama32}, Qwen3-4B-Base~\citep{qwen3technicalreport}, and Qwen2-1.5B-Instruct~\citep{yang2024qwen2}.
The first three are base checkpoints, whereas Qwen2-1.5B-Instruct has additionally been instruction-tuned.
These span two model families, parameter counts from 1B to 4B, and both base and instruction-tuned pretraining.

\paragraph{Benchmarks}
We use eight multiple-choice reasoning benchmarks: ARC Challenge and ARC Easy from the AI2 Reasoning Challenge~\citep{clark2018think}, BoolQ~\citep{clark2019boolq}, HellaSwag~\citep{zellers2019hellaswag}, OpenBookQA~\citep{mihaylov2018openbookqa}, PIQA~\citep{bisk2020piqa}, Social IQA~\citep{sap2019socialiqa}, and WinoGrande~\citep{sakaguchi2020winogrande}.
These vary in reasoning type and in the number of candidate answers, from two to four.
Our data preprocessing builds on the commonsense reasoning pipeline released with LLM-Adapters~\citep{hu-etal-2023-llm}: training uses the $n=170{,}300$ training examples pooled across the eight benchmarks, and evaluation uses their $22{,}419$ test examples.
In the main text we report uniformly-weighted average performance over the eight benchmarks; per-benchmark results are given in \Cref{app:full-results-backbones}.

\paragraph{Methods Compared}
For each value of $\beta$, the $m=8$ particles are trained jointly by minimising
\begin{equation}
R_\beta(Q_m) + \frac{\lambda_0}{2mn}\sum_{j=1}^m\|\bm{\theta}_j\|_2^2 ,
\label{eq:cs2-objective}
\end{equation}
where the ridge penalty, with $\lambda_0=0.01$, is commonly employed when training the parameters of a \ac{LoRA}.
We use the noisy AdamW update of \citet{nitanda2025propagation} and \citet{wang2026distributional}, which applies an AdamW step with learning rate $\eta=10^{-4}$ to \eqref{eq:cs2-objective} and then adds Gaussian noise with standard deviation $\sqrt{2\eta/(mn)}$ to each parameter (\Cref{app:implemented-update}).
All models are trained for three epochs, and results are reported at the final checkpoint unless otherwise stated.

\subsubsection{Pathologies of the Mixture Risk}
\label{sec:puzzle}
\label{subsec:exp-puzzle}

First we demonstrate that the failure of particle-based optimisation on the mixture risk does indeed occur in the aforementioned context.

\paragraph{Risk on the Training Data (H1)}

\begin{table}[!tbp]
\centering
\small
\setlength{\tabcolsep}{6pt}
\caption{Failure of particle-based optimisation fine-tuning an ensemble of \acp{LLM}. Training-set $R_{\mathrm{mix}}$ and $R_{\mathrm{ind}}$ for Qwen3-4B after three epochs of minimising \eqref{eq:cs2-objective}, on all $170{,}300$ optimisation examples. Pure mixture ($\beta=0$) attains the \emph{worst} $R_{\mathrm{mix}}$, despite the objective \eqref{eq:cs2-objective} being most closely-aligned to $R_{\mathrm{mix}}$. 
The higher value at $\beta=0.3$ than at $\beta=0.2$ is due to one particle that fits the training data poorly (\Cref{app:training-audit}).}
\label{tab:pro_optimisation_puzzle}
\begin{tabular}{lrr}
\toprule
Method & $R_{\mathrm{mix}}(Q_m)$ & $R_{\mathrm{ind}}(Q_m)$ \\
\midrule
Pure mixture, $\beta=0$      & $0.2610$ & $4.3658$ \\
Bridge, $\beta=0.01$ & $0.1566$ & $0.3840$ \\
Bridge, $\beta=0.1$ & $0.0199$ & $0.0607$ \\
Bridge, $\beta=0.2$ & $0.0101$ & $0.0253$ \\
Bridge, $\beta=0.3$ & $0.0835$ & $0.1332$ \\
Independent training, $\beta=1$     & $\mathbf{0.0039}$ & $\mathbf{0.0051}$ \\
\bottomrule
\end{tabular}
\end{table}

\Cref{tab:pro_optimisation_puzzle} reports the values of $R_{\mathrm{mix}}(Q_m)$ attained after ensemble-based fine-tuning of Qwen3-4B, computed based on the $170{,}300$ examples that comprise the training dataset.
Here, pure mixture training (i.e. $\beta = 0$) attains $R_{\mathrm{mix}}(Q_m)=0.2610$, thirteen times worse than the bridge at $\beta=0.1$ ($0.0199$), and worse even than the value $0.0051$ of $R_{\mathrm{ind}}(Q_m)$ attained by independent training.
The latter is surprising since, from Jensen's inequality, $R_{\mathrm{mix}} \leq R_{\mathrm{ind}}$.
The particle-based optimisation of $R_{\mathrm{mix}}$ is therefore failing in this context.

\begin{figure}[!tbp]
\centering
\includegraphics[width=\textwidth]{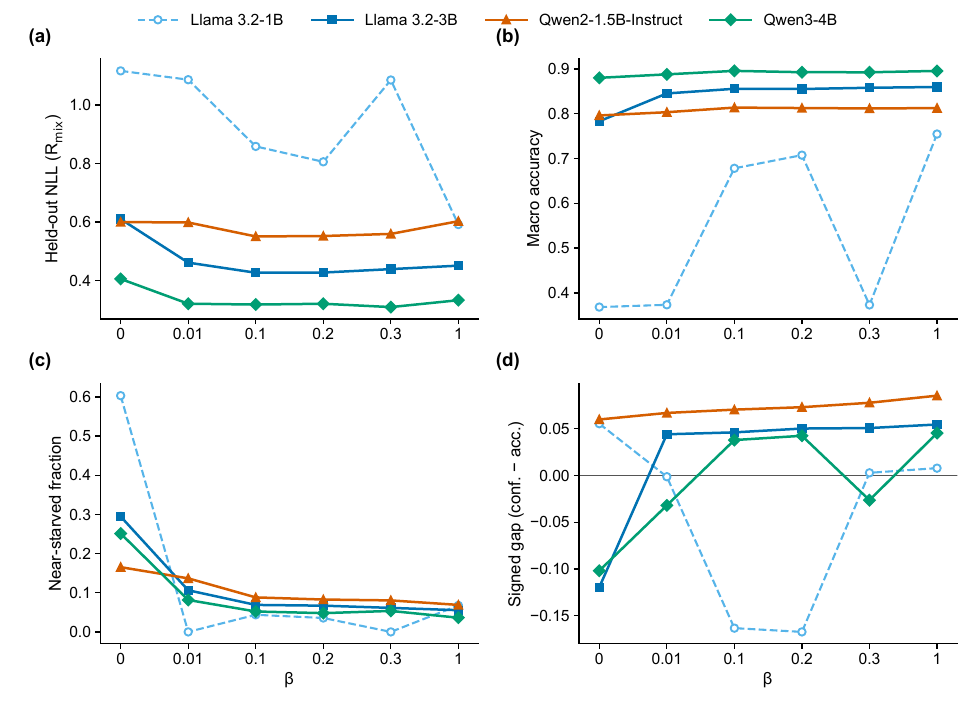}
\caption{Held-out results against $\beta$ for the four backbones, averaged over the eight test sets. (a) \ac{NLL}, which is the held-out $R_{\mathrm{mix}}$. (b) Accuracy. (c) Fraction of example and particle pairs with raw responsibility below $0.01$. (d) Signed gap, mean confidence minus accuracy. The six values of $\beta$ are equally spaced on the horizontal axis, and each setting was trained once.}
\label{fig:beta-sweep}
\end{figure}

\paragraph{Risk on the Held-Out Data}
The same picture holds out of sample (\Cref{fig:beta-sweep} and \Cref{tab:main-results} in \Cref{app:full-results-backbones}).
On every backbone the lowest held-out $R_{\mathrm{mix}}$ (i.e. the held-out \ac{NLL}) is attained with $\beta>0$, and pure mixture is the least accurate setting on all four backbones and has the worst held-out $R_{\mathrm{mix}}$ on three of them (e.g. $0.6104$, against $0.4271$ at $\beta=0.1$, on Llama 3.2-3B).
On all backbones but Llama 3.2-1B, most of the reduction in near-starved particle--example pairs is already obtained at small $\beta$, with little change beyond $\beta=0.1$.
On three of the four backbones the lowest held-out $R_{\mathrm{mix}}$ is attained at an interior $\beta\in(0,1)$, suggesting that the individual risk also acts as a regulariser that improves generalisation.

\paragraph{Calibration and Robustness}

Pure mixture is also markedly \emph{underconfident} on Llama 3.2-3B and Qwen3-4B, with signed gaps of $-0.12$ and $-0.10$, whereas every setting with $\beta>0$ has a signed gap within $0.06$ of zero (see also \Cref{fig:reliability}).
Since the mixture risk is a strictly proper scoring rule, this reflects a failure to attain a low value of $R_{\mathrm{mix}}$ rather than a property of the mixture risk itself.
Selecting the checkpoint with the lowest validation \ac{NLL} instead of the final one does not rescue pure mixture (\Cref{app:checkpoint-sensitivity}), although the ranking among positive values of $\beta$ depends on this choice, so the evidence does not single out a best positive $\beta$.

\subsubsection{Responsibility Starvation and Escape}
\label{subsec:exp-mechanism}

\begin{figure*}[!tbp]
  \centering
  \begin{subfigure}[t]{0.485\textwidth}
    \vspace{0pt}
    \centering
    \includegraphics[width=0.88\linewidth]{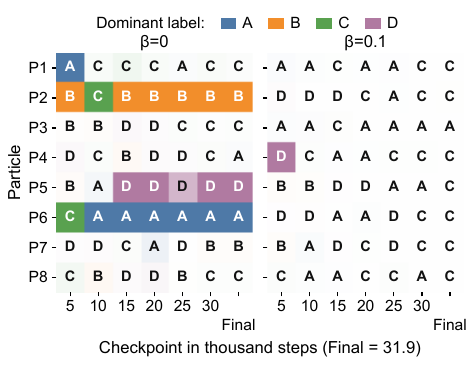}
    \caption{Label occupancy on HellaSwag.}
    \label{fig:mechanism-occupancy}
  \end{subfigure}\hfill%
  \begin{subfigure}[t]{0.485\textwidth}
    \vspace{0pt}
    \centering
    \includegraphics[width=0.88\linewidth]{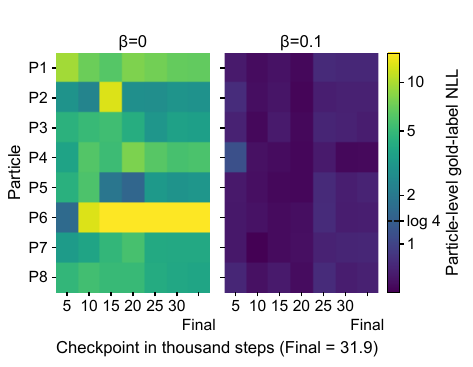}
    \caption{Particle-level gold-label \ac{NLL}.}
    \label{fig:mechanism-nll}
  \end{subfigure}
  \par\vspace{0.6em}
  \begin{subfigure}[t]{0.485\textwidth}
    \vspace{0pt}
    \centering
    \includegraphics[width=0.88\linewidth]{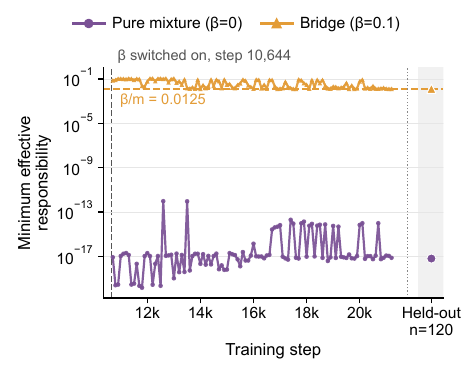}
    \caption{Minimum effective responsibility.}
    \label{fig:mechanism-floor}
  \end{subfigure}\hfill%
  \begin{subfigure}[t]{0.485\textwidth}
    \vspace{0pt}
    \centering
    \includegraphics[width=0.88\linewidth]{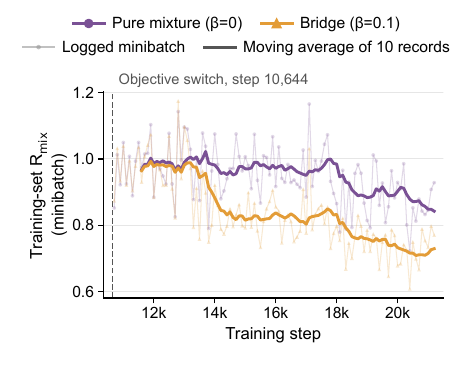}
    \caption{Training-set $R_{\mathrm{mix}}$.}
    \label{fig:mechanism-objective}
  \end{subfigure}
  \caption{Responsibility starvation and escape for Llama 3.2-3B. \emph{Top:} post hoc diagnostics at saved checkpoints for pure mixture ($\beta=0$, left of each panel) and the bridge ($\beta=0.1$, right). (a) Most frequent label per particle on HellaSwag. Under pure mixture three particles almost always predict one label, as in a slot partition (\Cref{def: slot partition}). (b) Particle-level \ac{NLL}. Under pure mixture every particle is worse than uniform guessing (\ac{NLL} at most $\log 4$), as part 1 of \Cref{thm:slot-degeneracy} predicts. \emph{Bottom:} training continued from the pure-mixture particles after one epoch with $\beta=0$ or $\beta=0.1$. (c) The minimum effective responsibility stays above $\beta/m=0.0125$ (\Cref{prop:responsibility-barrier}) under $\beta=0.1$ and falls to about $10^{-17}$ under $\beta=0$. (d) Training-set $R_{\mathrm{mix}}$ on minibatches is lower under $\beta=0.1$, although only $\beta=0$ optimises it (part 2 of \Cref{thm:slot-degeneracy}). More panels in \Cref{fig:starvation_dynamics_llama3b,fig:objective_switch_rescue}.
  }
  \label{fig:mechanism}
\end{figure*}

We now examine the particles of Llama 3.2-3B directly, with corroborating results for Qwen3-4B (\Cref{tab:particle_slot_diagnostics} and \Cref{app:additional-mechanism}).
The diagnostics we report are the raw responsibility \ac{ESS} $(\sum_j r_{ij}^2)^{-1}$, which equals $m=8$ under uniform responsibilities; the fraction of \emph{near-starved} particle--example pairs, with $r_{ij}<0.01$; the particle--answer mutual information and label occupancy, which measure how far the label predicted by a particle is determined by its identity, the defining feature of a slot partition; the spread of the particle-level \ac{NLL}; and the minimum effective responsibility.
The top row of \Cref{fig:mechanism} addresses (H2) and the bottom row addresses (H3).
The appendix gives the remaining panels of each experiment: \Cref{fig:starvation_dynamics_llama3b} adds the raw responsibility ESS and the mean responsibility of each particle for (H2), and \Cref{fig:objective_switch_rescue} adds the effective responsibility ESS and the spread of the particle-level \ac{NLL} for (H3).

\begin{table}[!tbp]
  \centering
  \caption{Post hoc particle diagnostics, averaged over test sets with equal weight; Llama 3.2-3B uses BoolQ, HellaSwag and OpenBookQA, and Qwen3-4B uses all eight test sets. MI is the particle--answer mutual information in nats. Occupancy is the fraction of examples on which a particle predicts its most frequent label, averaged (mean) or maximised (max) over particles. \ac{NLL} range is the spread of the particle-level \ac{NLL}.}
  \label{tab:particle_slot_diagnostics}
  \small
  \setlength{\tabcolsep}{3pt}
  \begin{tabular}{@{}llcccccccc@{}}
    \toprule
    & & & Raw & & MI & \multicolumn{2}{c}{Occupancy} & \ac{NLL} \\
    \cmidrule(lr){7-8}
    Backbone & Method & Checkpoint & ESS & $\Pr(r<0.01)$ & (nats) & Mean & Max & range \\
    \midrule
    Llama 3.2-3B & Pure mixture, $\beta=0$ & 5k & 4.506 & 0.371 & 0.410 & 0.702 & 1.000 & 8.724 \\
     & Bridge, $\beta=0.1$ & 5k & 6.626 & 0.034 & 0.013 & 0.544 & 1.000 & 0.689 \\
     & Pure mixture, $\beta=0$ & Final & 4.908 & 0.318 & 0.316 & 0.669 & 1.000 & 13.304 \\
     & Bridge, $\beta=0.1$ & Final & 6.978 & 0.061 & ${<}0.001$ & 0.397 & 0.429 & 0.200 \\
    \midrule
    Qwen3-4B & Pure mixture, $\beta=0$ & Final & 5.897 & 0.251 & 0.266 & 0.548 & 1.000 & 16.712 \\
     & Bridge, $\beta=0.1$ & Final & 7.243 & 0.052 & ${<}0.001$ & 0.383 & 0.390 & 0.079 \\
    \bottomrule
  \end{tabular}
\end{table}

\paragraph{Responsibility Starvation (H2)}

\begin{figure}[!tbp]
\centering
\includegraphics[width=\textwidth]{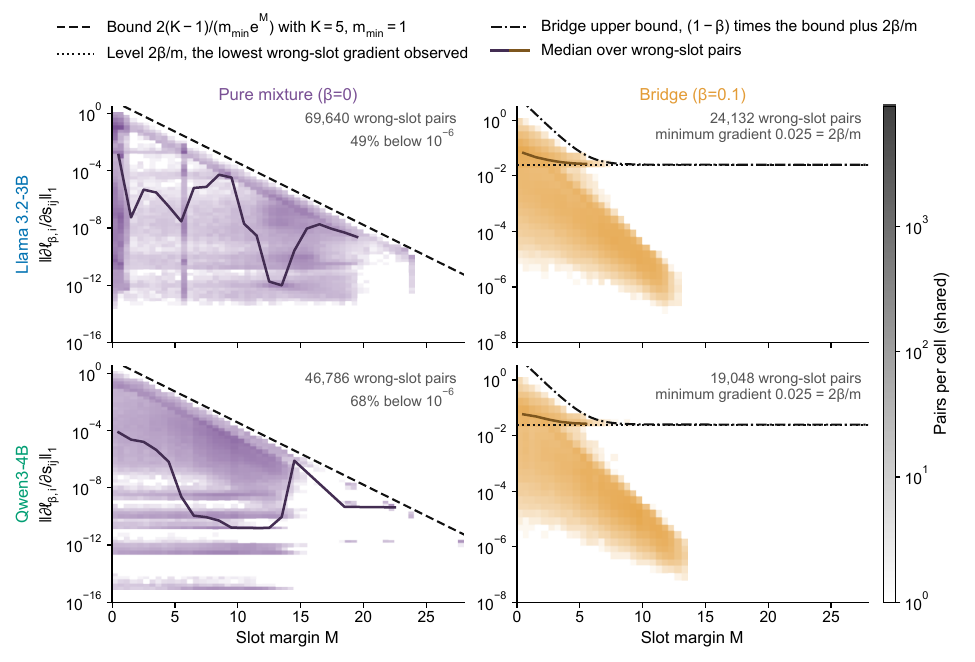}
\caption{Gradient received by each particle against its slot margin $M$ on the eight test sets, for Llama 3.2-3B (top) and Qwen3-4B (bottom), under pure mixture (left) and the bridge with $\beta=0.1$ (right). The vertical axis is the score-gradient norm $\|\nabla_{\bm{s}_{ij}}\ell_{\beta,i}\|_1$, and shading counts pairs of example and particle on a shared log scale. Dashed: the vanishing-gradient bound $2(K-1)/(m_{\min}e^{M})$ with $K=5$ and $m_{\min}=1$. Dash-dotted: $(1-\beta)$ times this bound plus $2\beta/m$. Dotted: $2\beta/m=0.025$. Solid: median over pairs whose particle's slot is a wrong label. Here $M$ is the smallest slot margin among particle $j$ and the particles whose slot is the correct label, and examples in which no particle's slot is the correct label are left out.}
\label{fig:prop4-test}
\end{figure}

Under optimisation of the mixture risk the particles drift towards a slot partition (\Cref{fig:mechanism-occupancy,fig:mechanism-nll} and \Cref{tab:particle_slot_diagnostics}).
The raw responsibility ESS stays between $4.5$ and $4.9$ out of $8$, against $6.6$ to $7.1$ under $\beta=0.1$; at the final checkpoint $31.8\%$ of pairs are near-starved, against $6.1\%$; the particle--answer mutual information is $0.3$ to $0.4$ nats, against nearly zero; and three of the eight particles predict a single label on every HellaSwag example.
As part 1 of \Cref{thm:slot-degeneracy} predicts, starved particles are not corrected: every particle is individually worse than uniform guessing throughout training, even though their mixture predicts well, whereas under $\beta=0.1$ every particle is a reasonable predictor.
The same pattern holds for Qwen3-4B.
Small responsibilities alone would not indicate a failure, since they can also occur at a good fit; here they coincide with poor particles and a poor training-set $R_{\mathrm{mix}}$.
At the level of gradients (\Cref{fig:prop4-test}), $49\%$ (Llama 3.2-3B) and $68\%$ (Qwen3-4B) of the pairs in which the particle's slot is a wrong label receive a gradient norm below $10^{-6}$ under the pure mixture risk, whereas under the bridge every such pair receives at least $2\beta/m=0.025$, as part 2 of \Cref{thm:slot-degeneracy} predicts.

\paragraph{Escape via $\beta > 0$ (H3)}
Continuing to train the pure-mixture particles after one epoch with $\beta=0.1$, rather than $\beta=0$, restores the training signal (\Cref{fig:mechanism-floor,fig:mechanism-objective,fig:objective_switch_rescue}).
The minimum effective responsibility stays above the floor $\beta/m$, against $10^{-20}$ to $10^{-12}$ under the control arm, the standard deviation of the particle-level \ac{NLL} collapses to below $0.5$ nats within about $20$ steps of the switch, and the training-set $R_{\mathrm{mix}}$ falls below that of the control arm, although only the control arm attempts to directly optimise it.
This is a finite-time counterpart of part 2 of \Cref{thm:slot-degeneracy}, observed along a single trajectory per arm.

\begin{figure}[!tbp]
\centering
\includegraphics[width=\textwidth]{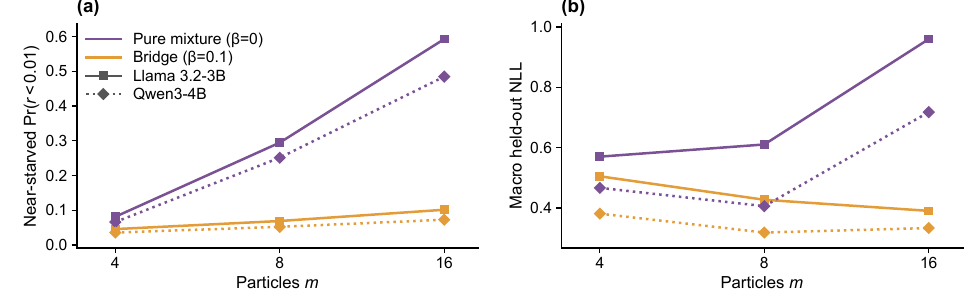}
\caption{Varying the number of particles $m$ for Llama 3.2-3B (squares, solid) and Qwen3-4B (diamonds, dotted) under pure mixture and the bridge with $\beta=0.1$. (a) Fraction of example and particle pairs with raw responsibility below $0.01$, averaged over the eight test sets. (b) Held-out \ac{NLL}, averaged over the eight test sets. }
\label{fig:m-ablation}
\end{figure}

\paragraph{Number of Particles}
More particles make starvation worse under the pure mixture risk, but not under the bridge (\Cref{fig:m-ablation} and \Cref{tab:m_ablation}).
From $m=8$ to $m=16$, the near-starved fraction under pure mixture roughly doubles and its held-out $R_{\mathrm{mix}}$ rises from $0.6104$ to $0.9596$ on Llama 3.2-3B and from $0.4059$ to $0.7177$ on Qwen3-4B, whereas under $\beta=0.1$ it changes little ($0.4271$ to $0.3904$, and $0.3185$ to $0.3335$).

\subsubsection{The Role of Backbone Competence (H4)}
\label{subsec:exp-competence}
\label{sec:cross-backbone}

The implicit responsibility floor of \Cref{cor:competence-floor} should make $\beta>0$ matter less when the backbone \ac{LLM} is more competent.
Using the macro accuracy of pure mixture as a proxy for competence (\Cref{tab:cross-backbone-summary}), this is what we observe.
For each backbone we report the smallest $\beta$ at which the macro accuracy comes within $0.01$ of that of independent training (\Cref{tab:main-results}).
Llama 3.2-1B is the least competent backbone, with a pure-mixture accuracy of $0.368$ that is the same as its zero-shot accuracy, and needed $\beta=1$, since every smaller $\beta$ leaves it at least $0.047$ below independent training.
Llama 3.2-3B is the second least competent ($0.783$) and needed $\beta=0.1$.
Qwen2-1.5B-Instruct ($0.796$) and Qwen3-4B ($0.880$) are the two most competent and needed only $\beta=0.01$.
The more competent the backbone, the smaller the $\beta$ that it needs, and thus we see a trend consistent with (H4).
Measuring competence before fine-tuning, as in \Cref{cor:competence-floor}, orders the two model families in the same way: the mean correct-label probability of the frozen backbone is $0.40$ for both Llama models, against $0.50$ and $0.57$ for the two Qwen models (\Cref{tab:pre_adaptation,fig:competence}).

\begin{table}[!tbp]
\centering
\caption{Cross-backbone summary. 
Accuracy (top) and \ac{NLL} (bottom) of the frozen backbone (zero-shot) and after fine-tuning under pure mixture and under the best bridge setting for that metric. Backbones are ordered by pure-mixture accuracy, our proxy for base competence.
}

\label{tab:cross-backbone-summary}
\small
\setlength{\tabcolsep}{5pt}
\begin{tabular}{lccc}
\toprule
Backbone & Zero-shot & Pure mixture & Best bridge \\
\midrule
\multicolumn{4}{l}{\emph{Accuracy $\uparrow$}} \\
Llama 3.2-1B & 0.3681 & 0.3683 & 0.7076 ($\beta=0.2$) \\
Llama 3.2-3B & 0.3697 & 0.7832 & 0.8580 ($\beta=0.3$) \\
Qwen2-1.5B-Instruct & 0.5696 & 0.7962 & 0.8138 ($\beta=0.1$) \\
Qwen3-4B & 0.6932 & 0.8802 & 0.8958 ($\beta=0.1$) \\
\midrule
\multicolumn{4}{l}{\emph{\ac{NLL} $\downarrow$}} \\
Llama 3.2-1B & 1.3884 & 1.1162 & 0.8058 ($\beta=0.2$) \\
Llama 3.2-3B & 1.5594 & 0.6104 & 0.4271 ($\beta=0.1$) \\
Qwen2-1.5B-Instruct & 0.9500 & 0.6001 & 0.5510 ($\beta=0.1$) \\
Qwen3-4B & 0.6858 & 0.4059 & 0.3099 ($\beta=0.3$) \\
\bottomrule
\end{tabular}
\end{table}

\subsubsection{Comparison with \texorpdfstring{$\alpha$}{alpha}-R\'{e}nyi Training}
\label{subsec:exp-renyi}

Finally, we compare with the $\alpha$-R\'{e}nyi approach of \citet{cordero2026soft}, implemented on Llama 3.2-3B with $m=8$ and their principal setting $\alpha=0.8$, both with and without injected noise (\Cref{app:implemented-update}).
Their objective is defined in sequence space, rather than ours which is defined across the multiple-choice answer set; we therefore evaluate each approach with the predictive that it optimises to ensure an impartial assessment.
The bridge with $\beta=0.1$ matches the $\alpha$-R\'{e}nyi approach in accuracy ($0.8558$, against $0.8529$ and $0.8508$) and improves \ac{NLL}, ECE and Brier score (\Cref{tab:renyi-native-comparison}).
Since the predictive distributions differ as well as the interpolation paths (cf. \Cref{rem: reyni}), this is a comparison of methods rather than of interpolation paths; implementation details are given in \Cref{app:sequence-space-full}.

\begin{table}[!tbp]
\centering
\small
\setlength{\tabcolsep}{5pt}
\caption{Comparison with $\alpha$-R\'{e}nyi regularisation on Llama 3.2-3B after three epochs. 
Each objective is evaluated with the predictive distribution that it optimises, with performance averaged over each of the eight benchmarks considered.
\emph{With noise} uses the noisy AdamW update employed for the bridge (\Cref{app:implemented-update}), so that the two methods differ only in their objective; \emph{without noise} omits the injected Gaussian noise, as in \citet{cordero2026soft}.
}
\label{tab:renyi-native-comparison}
\begin{tabular}{lcccc}
\toprule
Method
& Accuracy \(\uparrow\)
& \ac{NLL} \(\downarrow\)
& ECE \(\downarrow\)
& Brier \(\downarrow\)
\\
\midrule
Sequence \(\alpha=0.8\), without noise
& 0.8529 & 0.4862 & 0.0545 & 0.2232
\\
Sequence \(\alpha=0.8\), with noise
& 0.8508 & 0.4854 & 0.0537 & 0.2215
\\
Bridge, \(\beta=0.1\)
& \textbf{0.8558}
& \textbf{0.4271}
& \textbf{0.0508}
& \textbf{0.2163}
\\
\bottomrule
\end{tabular}
\end{table}

\FloatBarrier

\section{Discussion}
\label{sec:discussion}

This paper set out to explain why gradient-based minimisation of the mixture risk over a finite set of particles has been observed to perform poorly, both when computing \ac{PrO} posteriors and when jointly training ensembles.
Although $R_{\mathrm{mix}}$ is convex in $Q$, this convexity is lost when $Q$ is restricted to a particle approximation $Q_m$: the mixture risk constrains only the aggregate predictive mixture, not how it is decomposed across particles, so that a particle whose responsibilities vanish is starved of training signal.
We proved that, as the particles approach a slot partition, the pure-mixture gradient that would correct a wrongly-specialised particle vanishes (\Cref{thm:slot-degeneracy}), whereas the bridge adds a reverse-\ac{KL} barrier on the responsibilities (\Cref{prop:responsibility-barrier}), so that every effective responsibility is at least $\beta/m$ and a slot partition is escaped in finite time.
At the distributional level, the bridge regularises the \ac{PrO} posterior towards a $\beta$-tempered Bayesian posterior (\Cref{prop:tempered-reference}).

The two case studies support this account.
In Case Study~1, \ac{MFLD} applied to the mixture risk failed to converge, even with $m=128$ particles and $10^{7}$ iterations, whereas with $\beta=0.01$ every chain converged to a common solution, which was also the best calibrated.
In Case Study~2, pure mixture training attained a training-set value of $R_{\mathrm{mix}}$ larger than the value of $R_{\mathrm{ind}}$ attained by independent training (\Cref{tab:pro_optimisation_puzzle}); since $R_{\mathrm{mix}}\leq R_{\mathrm{ind}}$ by Jensen's inequality, this is direct evidence of a failure of optimisation.
A small $\beta>0$ improved the held-out value of $R_{\mathrm{mix}}$ on every backbone; particle-level diagnostics revealed a drift towards slot partitions under pure mixture, which $\beta>0$ prevented and, when switched on part-way through training, reversed; and the benefit of $\beta>0$ decreased as the competence of the backbone increased, as anticipated by \Cref{cor:competence-floor}.

Several limitations should be noted.
First, our analysis concerns the training signal on the scores of individual examples as a slot partition is approached, rather than the parameters themselves, so it provides a phenomenological explanation rather than a global convergence result.
Second, Case Study~1 is a low-dimensional synthetic example, chosen because its \ac{PrO} posterior is known to differ materially from the Bayesian posterior; computation of \ac{PrO} posteriors for higher-dimensional models remains to be explored.
Third, in Case Study~2 each configuration was trained once, and model family, pretraining data and instruction tuning are confounding factors across the four backbones.
Fourth, our results do not single out a best positive value of $\beta$, since the ranking among positive values depends on the metric and on how the checkpoint is selected.
Finally, our comparison with the $\alpha$-R\'{e}nyi approach of \citet{cordero2026soft} differs in the predictive distribution as well as in the interpolation path.

For practitioners, our results suggest that a small positive $\beta$ is an inexpensive safeguard against responsibility starvation, and that the value of $\beta$ required decreases as the competence of the initialisation increases.
For \ac{PrO} posteriors, $Q_\beta^{(n)}$ with a small fixed $\beta$ can be reported as a regularised \ac{PrO} posterior; whether $\beta$ can instead be annealed to zero during computation, so that $Q_{\mathrm{PrO}}$ itself is targeted, is an interesting open question, as is the adaptive selection of $\beta$ based on the observed responsibilities.
Since the bridge retains a non-zero mixture term for every $\beta<1$, it preserves the joint training of particles that motivates both \ac{PrO} posteriors and the joint training of ensembles.

\paragraph{Acknowledgements}

YL was supported by a five-year PGTA scholarship from the School of Computing, Newcastle University.
CJO was supported by a Philip Leverhulme Prize (PLP-2023-004). DTF was supported by the Australian Research Council (DP250101069). 
This research was supported by the NVIDIA Academic Grant Program.
The authors are grateful to Francesca Crucinio for helpful discussion of related work.

\bibliographystyle{abbrvnat}
\bibliography{bibliography}

\input{appendix}

\end{document}

%% file: appendix.tex
\clearpage
\appendix
\section*{Appendices}

The appendices are organised as follows.
\Cref{app:theory} contains the proofs of the theoretical results stated in the main text.
\Cref{app:experimental-details} records the computational resources used in Case Study~2, and the implementation of the particle schemes used in Case Studies~1 and~2.
\Cref{app:empirical} contains additional empirical results, ordered according to where they are referred to in \Cref{sec: experiments}.

\section{Proofs}
\label{app:theory}

\subsection{Proof of \Cref{prop:responsibility-barrier}}
\label{app:proof-responsibility-barrier}

Recall that
\[
    w_{ij}
    =
    p_{\bm{\theta}_j}(y_i\mid x_i)
\]
denotes the probability assigned by particle \(j\) to the correct
answer for example \(i\). 
For the following proof it is convenient to use the notation
\begin{align}
    R_{\mathrm{mix}} = \frac{1}{n} \sum_{i=1}^n \ell_{0,i}, \qquad  \ell_{0,i} =
    -\log
    \left(
        \frac{1}{m}
        \sum_{j=1}^{m} w_{ij}
    \right),
    \label{eq:app-ell-zero}
    \\
    R_{\mathrm{ind}} = \frac{1}{n} \sum_{i=1}^n \ell_{\mathrm{ind},i}, \qquad  \ell_{\mathrm{ind},i} =
    -\frac{1}{m}
    \sum_{j=1}^{m}
    \log w_{ij},
    \label{eq:app-ell-bayes}
    \\
    R_\beta = \frac{1}{n} \sum_{i=1}^n \ell_{\beta,i}, \qquad \ell_{\beta,i} =
    (1-\beta)\ell_{0,i}
    +
    \beta\ell_{\mathrm{ind},i} .
    \label{eq:app-ell-beta}
\end{align}

\begin{proof}[Proof of Proposition~\ref{prop:responsibility-barrier}]
Using \(S_i=\sum_j w_{ij}\) and
\(r_{ij}=w_{ij}/S_i\), we obtain
\begin{align}
    \operatorname{KL}(u_m \Vert r_i)
    =
    \frac{1}{m}
    \sum_{j=1}^{m}
    \log
    \frac{1/m}{r_{ij}}
    & =
    \frac{1}{m}
    \sum_{j=1}^{m}
    \log
    \frac{S_i}{m w_{ij}}
    \nonumber\\
    &=
    \log\frac{S_i}{m}
    -
    \frac{1}{m}
    \sum_{j=1}^{m}
    \log w_{ij}
    =
    \ell_{\mathrm{ind},i}
    -
    \ell_{0,i}.
    \label{eq:app-kl-barrier}
\end{align}
Therefore,
\begin{align*}
    \ell_{\beta,i}
    =
    (1-\beta)\ell_{0,i}
    +
    \beta\ell_{\mathrm{ind},i}
    =
    \ell_{0,i}
    +
    \beta
    \operatorname{KL}(u_m \Vert r_i).
\end{align*}
as claimed.
\end{proof}

\subsection{Proof of \Cref{thm:slot-degeneracy}}
\label{app: proof gradients}

For a training example $i$ we abbreviate $\mathcal{S}_i=S_{y_i}(x_i)$ and $m_i=m_{y_i}(x_i)$, the set and number of particles whose slot at $x_i$ is the correct label $y_i$.

\subsubsection{Responsibilities along the Slot-Margin Family}
\label{app:proof-slot-fingerprints}

Recall that, along the slot-margin family \eqref{eq:slot-margin-main}, the correct-label probabilities $w_{ij}$ and raw responsibilities $r_{ij}$ for example $i$ depend only on the slots $c_1(x_i),\dots,c_m(x_i)$ and the label $y_i$. 
The proof of \Cref{thm:slot-degeneracy} requires only the following elementary calculation.

\begin{lemma}[Responsibilities along the slot-margin family]
\label{thm:slot-fingerprints}
Let $D_M=e^M+K-1$ and $A_i(M)=m_i e^M+m-m_i$. Along the slot-margin family, for each example $i$, particle $j$ assigns probability $e^M/D_M$ to its slot label $c_j(x_i)$ and probability $1/D_M$ to every other label, and
\begin{align}
  r_{ij}&=
  \begin{cases}
    e^M/A_i(M),&j\in \mathcal{S}_i,\\
    1/A_i(M),&j\notin \mathcal{S}_i.
  \end{cases}
  \label{eq:slot-r-fingerprint}
\end{align}
\end{lemma}

\begin{proof}[Proof of \Cref{thm:slot-fingerprints}]
Fix an example $i$ and write $y=y_i$.
Under the slot-margin construction \eqref{eq:slot-margin-main},
the softmax denominator in \eqref{eq:class-probability} is
\(
    D_M=e^M+K-1
\).
Particle \(j\) therefore assigns probability
\(
    e^M/D_M
\)
to its slot label \(c_j(x_i)\), and probability
\(
    1/D_M
\)
to every other label.
The total correct-label probability across particles is
\begin{align}
    \sum_{j=1}^{m}
    w_{ij}
    =
    m_i\frac{e^M}{D_M}
    +
    (m-m_i)\frac{1}{D_M}
    =
    \frac{A_i(M)}{D_M} ,
    \label{eq:app-total-correct-prob}
\end{align}
and dividing $w_{ij}$, which equals $e^M/D_M$ if $j \in \mathcal{S}_i$ and $1/D_M$ otherwise, by \eqref{eq:app-total-correct-prob} gives \eqref{eq:slot-r-fingerprint}.
\end{proof}

\begin{remark}[Diagnostic signatures of a slot partition]
\label{rem:slot-fingerprints}
Although not needed for the proof of \Cref{thm:slot-degeneracy}, the same calculation also identifies the behaviour of quantities that are used as diagnostics in \Cref{sec:experiments}.
Let
\(
\operatorname{ESS}_i=\bigl(\sum_{j=1}^m r_{ij}^2\bigr)^{-1}
\)
denote the effective sample size of the raw responsibilities for example $i$, which ranges from $1$ (all responsibility on a single particle) to $m$ (uniform responsibilities).
From \eqref{eq:app-total-correct-prob} and \eqref{eq:slot-r-fingerprint},
\begin{align*}
  \bar p(y_i\mid x_i)=\frac{A_i(M)}{mD_M},
  \qquad
  \operatorname{ESS}_i
  =\frac{A_i(M)^2}{m_i e^{2M}+m-m_i} ,
\end{align*}
and, applying the same calculation to an arbitrary label $c$ in place of $y_i$,
\begin{equation*}
\bar p(c\mid x_i) = \frac{m_c(x_i) e^M + m - m_c(x_i)}{m D_M} .
\end{equation*}
Consequently, if $m_i\geq1$, then as $M\to\infty$,
\begin{equation*}
  -\log \bar p(y_i\mid x_i)\to-\log\frac{m_i}{m},
  \qquad
  \operatorname{ESS}_i\to m_i,
  \qquad
  \max_{c\in\mathcal{C}} \bar p(c\mid x_i)\to\max_{c\in\mathcal{C}}\frac{m_c(x_i)}{m}.
\end{equation*}
In particular, close to a slot partition the raw-responsibility effective sample size for example $i$ is approximately the number of particles whose slot is the correct label, rather than $m$.
\end{remark}

\subsubsection{Main Argument}
\label{app:proof-slot-degeneracy}

\begin{proof}[Proof of \Cref{thm:slot-degeneracy}]

Our starting point is the expression
\(
\nabla_{\bm{s}_{ij}}\ell_{\beta,i}=\rho_{ij}^{(\beta)}(\bm{p}_{ij}-\bm{e}_{y})
\)
from \eqref{eq:score-gradient}, where at \(\beta=0\) the responsibility is \(\rho_{ij}^{(0)}=r_{ij}\).
\Cref{thm:slot-fingerprints} enters in two ways: its expression for the particle probabilities determines $\bm{p}_{ij}$, and its expression \eqref{eq:slot-r-fingerprint} determines $r_{ij}$.
For a particle \(j\in \mathcal{S}_i\), the vector $\bm{p}_{ij}$ has entry $e^M/D_M$ at the correct label $y$ and entries $1/D_M$ elsewhere, so that
\[
    \left\|
        \bm{p}_{ij}-\bm{e}_y
    \right\|_1
    =
    \left(1-\frac{e^M}{D_M}\right)+\frac{K-1}{D_M}
    =
    \frac{
        2(K-1)
    }{
        D_M
    }.
\]
Therefore, using $r_{ij} = e^M/A_i(M)$ from \eqref{eq:slot-r-fingerprint} together with $A_i(M) \geq m_i e^M$ and $D_M \geq e^M$,
\begin{align}
    \left\|
        \nabla_{\bm{s}_{ij}}
        \ell_{0,i}
    \right\|_1
    =
    r_{ij}
    \left\|
        \bm{p}_{ij}-\bm{e}_y
    \right\|_1
    =
    \frac{e^M}{A_i(M)}
    \frac{2(K-1)}{D_M}
    \leq
    \frac{
        2(K-1)
    }{
        m_i e^M
    }.
    \label{eq:app-matching-gradient-bound}
\end{align}
For a wrong-slot particle \(j\notin \mathcal{S}_i\), the vector $\bm{p}_{ij}$ has entry $e^M/D_M$ at $c_j(x_i) \neq y$ and entries $1/D_M$ elsewhere, so that
\[
    \left\|
        \bm{p}_{ij}-\bm{e}_y
    \right\|_1
    =
    2
    \left(
        1-\frac{1}{D_M}
    \right)
    \leq2.
\]
Consequently, using $r_{ij} = 1/A_i(M)$ from \eqref{eq:slot-r-fingerprint} and $A_i(M) \geq m_i e^M$,
\begin{align}
    \left\|
        \nabla_{\bm{s}_{ij}}
        \ell_{0,i}
    \right\|_1
    =
    \frac{1}{A_i(M)}
    \left\|
        \bm{p}_{ij}-\bm{e}_y
    \right\|_1
    \leq
    \frac{2}{m_i e^M}
    \leq
    \frac{
        2(K-1)
    }{
        m_i e^M
    },
    \label{eq:app-wrong-gradient-bound}
\end{align}
where the final inequality uses \(K\geq2\).
Since \(m_i\geq m_{\min}\), Eqs.~\eqref{eq:app-matching-gradient-bound}
and~\eqref{eq:app-wrong-gradient-bound} give
\[
    \max_{i,j}
    \left\|
        \nabla_{\bm{s}_{ij}}\ell_{0,i}
    \right\|_1
    \leq
    \frac{
        2(K-1)
    }{
        m_{\min}e^M
    }
    \longrightarrow0.
\]

On the other hand, for a fixed \(\beta>0\) and a wrong-slot particle \(j\notin \mathcal{S}_i\), \Cref{thm:slot-fingerprints} gives $r_{ij} = 1/A_i(M)$ and $\bm{p}_{ij} = (e^M \bm{e}_{c_j(x_i)} + \sum_{c \neq c_j(x_i)} \bm{e}_c)/D_M$, so that as $M \rightarrow \infty$,
\[
    r_{ij}\longrightarrow0,
    \qquad
    \rho_{ij}^{(\beta)}
    =(1-\beta)r_{ij}+\frac{\beta}{m}
    \longrightarrow
    \frac{\beta}{m},
    \qquad
    \bm{p}_{ij}
    \longrightarrow
    \bm{e}_{c_j(x_i)}.
\]
Thus,
\[
    \nabla_{\bm{s}_{ij}}\ell_{\beta,i}
    \longrightarrow
    \frac{\beta}{m}
    \left(
        \bm{e}_{c_j(x_i)}-\bm{e}_y
    \right),
\]
and because \(c_j(x_i)\neq y\),
\(
    \|
        \bm{e}_{c_j(x_i)}-\bm{e}_y
    \|_1
    =2
\).
Therefore,
\(
    \|
        \nabla_{\bm{s}_{ij}}\ell_{\beta,i}
    \|_1
    \to
    2\beta/m
\).

The final claim is proven as \Cref{thm:escape-times} in \Cref{app:proof-escape-times}.
\end{proof}

\subsubsection{Escape Times for the Bridge}
\label{app:proof-escape-times}

For a non-asymptotic statement, we say that the particles are in an \emph{$\varepsilon$-slot state} for example $i$ if each particle $j$ assigns probability at least $1-\varepsilon$ to its slot label $c_j(x_i)$, and at most $\varepsilon$ to every other label; i.e. $p_j(c_j(x_i)\mid x_i)\geq1-\varepsilon$ and $p_j(c\mid x_i)\leq\varepsilon$ for all $c\neq c_j(x_i)$.
Throughout we assume \(0<\varepsilon<1/2\).

\begin{proposition}[Escape Times]
\label{thm:escape-times}
Consider an example $i$ with $m_i\geq1$, and suppose the particles are in an $\varepsilon$-slot state for example $i$.
For a wrong-slot particle $j\notin \mathcal{S}_i$,
\begin{equation}
  r_{ij}\leq\frac{\varepsilon}{m_i(1-\varepsilon)}.
  \label{eq:slot-wrong-resp-upper}
\end{equation}
Moreover, for $\beta>0$, consider the gradient flow $\dot{\bm{s}}_{ij}=-\nabla_{\bm{s}_{ij}}\ell_{\beta,i}$ for the single example $i$, started from a wrong-slot margin
\begin{equation}
    M_j(t)
    =
    s_j(c_j(x_i)\mid x_i)-s_j(y_i\mid x_i)
    \label{eq:app-wrong-slot-margin}
\end{equation}
of magnitude $M_j(0)=M_0>0$.
Then, for as long as $p_j(c_j(x_i)\mid x_i)\geq1-\varepsilon$ and $p_j(y_i\mid x_i)\leq\varepsilon$, the margin satisfies $\dot M_j(t)\leq-2\beta(1-\varepsilon)/m$, and particle $j$ leaves this region within time
\begin{equation}
  T_{\mathrm{esc}}
  \leq\frac{mM_0}{2\beta(1-\varepsilon)}.
  \label{eq:slot-escape-time}
\end{equation}
\end{proposition}

\begin{proof}[Proof of Theorem~\ref{thm:escape-times}]
Write $y=y_i$.
By the definition of an
\(\varepsilon\)-slot state, a wrong-slot particle \(j\notin \mathcal{S}_i\) has
\(
    w_{ij}
    =
    p_j(y\mid x_i)
    \leq\varepsilon
\),
while every particle \(k\in \mathcal{S}_i\) has
\(
    w_{ik}
    =
    p_k(y\mid x_i)
    \geq1-\varepsilon
\).
Therefore
\(
    \sum_{k=1}^{m}w_{ik}
    \geq
    m_i(1-\varepsilon)
\),
and hence
\begin{equation}
    r_{ij}
    =
    \frac{w_{ij}}
         {\sum_k w_{ik}}
    \leq
    \frac{
        \varepsilon
    }{
        m_i(1-\varepsilon)
    }.
    \label{eq:app-epsilon-responsibility}
\end{equation}

We now prove the escape-time statement.
Write \(p_{j,c}=p_j(c\mid x_i)\) and \(c^\ast=c_j(x_i)\) for brevity.
From \eqref{eq:score-gradient}, the \(c\)th component of \(\dot{\bm{s}}_{ij}=-\nabla_{\bm{s}_{ij}}\ell_{\beta,i}\) is \(-\rho_{ij}^{(\beta)}(p_{j,c}-\mathbf{1}\{c=y\})\), so that
\begin{align}
    \dot M_j(t)
    =
    -
    \rho_{ij}^{(\beta)}
    \left[
        p_{j,c^\ast}
        -
        \bigl(p_{j,y}-1\bigr)
    \right]
    =
    -
    \rho_{ij}^{(\beta)}
    \left[
        p_{j,c^\ast}
        +
        1-p_{j,y}
    \right].
    \label{eq:app-margin-derivative}
\end{align}
While
\(
    p_{j,c^\ast}\geq1-\varepsilon
\)
and
\(
    p_{j,y}\leq\varepsilon
\),
and since
\(
    \rho_{ij}^{(\beta)}
    \geq
    \beta/m
\)
always,
\begin{equation}
    \dot M_j(t)
    \leq
    -
    \frac{
        2\beta(1-\varepsilon)
    }{
        m
    }.
    \label{eq:app-margin-drift}
\end{equation}
Integrating \eqref{eq:app-margin-drift} gives
\[
    M_j(t)
    \leq
    M_0
    -
    \frac{
        2\beta(1-\varepsilon)
    }{
        m
    }t
\]
for as long as the particle remains in the region.
However, within the region \(p_{j,c^\ast}\geq1-\varepsilon>\varepsilon\geq p_{j,y}\), so that \(M_j(t)>0\) there.
It follows that the particle must leave the region before the right-hand side reaches zero, i.e. within time
\(
    T_{\mathrm{esc}}
    \leq
    mM_0/(2\beta(1-\varepsilon))
\).
\end{proof}

\subsection{Proof of \Cref{cor:competence-floor}}
\label{app:proof-competence-floor}

\begin{proof}[Proof of \Cref{cor:competence-floor}]
For every particle--example pair satisfying \(w_{ij}\geq\kappa\),
\[
    \sum_{k=1}^{m}w_{ik}
    \leq m,
\]
because every \(w_{ik}\leq1\). Therefore,
\[
    r_{ij}
    =
    \frac{
        w_{ij}
    }{
        \sum_{k=1}^{m}w_{ik}
    }
    \geq
    \frac{\kappa}{m}.
\]
It follows that
\begin{align*}
    \rho_{ij}^{(\beta)}
    =
    (1-\beta)r_{ij}
    +
    \frac{\beta}{m}
    \geq
    (1-\beta)\frac{\kappa}{m}
    +
    \frac{\beta}{m}
    =
    \frac{
        \beta+(1-\beta)\kappa
    }{
        m
    }.
\end{align*}
\end{proof}

\subsection{Proof of \Cref{prop:tempered-reference}}
\label{app:proof-tempered-reference}

\begin{proof}[Proof of Proposition~\ref{prop:tempered-reference}]

By the definition of \(\Pi_{n,\beta}\),
\begin{equation}
\log
\frac{\mathrm d\Pi_{n,\beta}}
     {\mathrm d\Pi}(\theta)
=
\beta
\sum_{i=1}^{n}
\log p_{\theta}(y_i\mid x_i)
-
\log Z_{n,\beta}.
\label{eq:app-tempered-log-density}
\end{equation}
Therefore,
\begin{align}
\operatorname{KL}
\left(
Q\Vert\Pi_{n,\beta}
\right)
&=
\int
\log
\frac{\mathrm dQ}
     {\mathrm d\Pi_{n,\beta}}(\theta)
\,\mathrm dQ(\theta)
\nonumber\\
&=
\int
\log
\frac{\mathrm dQ}
     {\mathrm d\Pi}(\theta)
\,\mathrm dQ(\theta)
-
\int
\log
\frac{\mathrm d\Pi_{n,\beta}}
     {\mathrm d\Pi}(\theta)
\,\mathrm dQ(\theta)
\nonumber\\
&=
\operatorname{KL}(Q\Vert\Pi)
-
\beta
\sum_{i=1}^{n}
\int
\log p_{\theta}(y_i\mid x_i)
\,\mathrm dQ(\theta)
+
\log Z_{n,\beta}
\nonumber\\
&=
\operatorname{KL}(Q\Vert\Pi)
+
n\beta
R_{\mathrm{ind}}(Q)
+
\log Z_{n,\beta}.
\label{eq:app-tempered-kl-expansion}
\end{align}
Rearranging gives
\begin{equation}
\beta
R_{\mathrm{ind}}(Q)
+
\frac{1}{n}
\operatorname{KL}(Q\Vert\Pi)
=
\frac{1}{n}
\operatorname{KL}
\left(
Q\Vert\Pi_{n,\beta}
\right)
-
\frac{1}{n}
\log Z_{n,\beta}.
\label{eq:app-tempered-rearrangement}
\end{equation}
Substituting
\eqref{eq:app-tempered-rearrangement}
into
\eqref{eq:bridge-objective}
yields the objective functional
\begin{equation}
J_\beta(Q)
=
(1-\beta)
R_{\mathrm{mix}}(Q)
+
\frac{1}{n}
\operatorname{KL}
\left(
Q\Vert\Pi_{n,\beta}
\right)
-
\frac{1}{n}
\log Z_{n,\beta} ,
\end{equation}
which establishes the result, since \(\log Z_{n,\beta}\) is a constant in \(Q\).
\end{proof}

\section{Experimental Details}
\label{app:experimental-details}

This appendix records the computational resources used in Case Study~2 (\Cref{app:computational_resources}), and the implementation of the particle schemes used in both case studies (\Cref{app:mfld-derivation}).

\subsection{Computational Resources}
\label{app:computational_resources}

All experiments in Case Study~2 were run on a single node with eight NVIDIA H100 80GB SXM GPUs, using PyTorch 2.5.1 with CUDA 12.1, Transformers 5.9.0 and Accelerate 1.13.0; \acp{LoRA} are implemented directly in the training code, and the backbone is evaluated in bfloat16.
Each \ac{LoRA} has rank $32$, scaling factor $2$ and dropout $0.05$, and is attached to the query, key and value projections of every attention block and to the up and down projections of every feed-forward block, with $\bm{A}$ initialised by the Kaiming uniform scheme and $\bm{B}$ at zero.
Runs with $m\in\{4,8\}$ use four GPUs and runs with $m=16$ use eight, with the particles divided equally between the GPUs and the frozen backbone shared by the particles on each GPU.
All runs except the objective-switch experiment use three epochs ($31{,}932$ optimisation steps), a batch size of $16$ and a microbatch size of $8$, with particle seeds $42$ to $41+m$ and checkpoints saved every $5{,}000$ steps.
Since each particle requires only the correct-label probabilities of the other particles, joint training costs about the same as independent training (\Cref{tab:computational_resources}).
In total the experiments took $1{,}082$ GPU hours, of which $544$ were for the $24$ runs in \Cref{tab:main-results}.

\begin{table}[!tbp]
\centering
\caption{Cost of joint training with $m=8$ particles for three epochs. Wall time and peak device memory per GPU are for independent training ($\beta=1$). The relative columns give the range over $\beta\in\{0,0.01,0.1,0.2,0.3\}$ of the training time and peak memory, divided by those of independent training on the same backbone.}
\label{tab:computational_resources}
\small
\begin{tabular}{@{}lcccc@{}}
\toprule
& \multicolumn{2}{c}{Independent training} & \multicolumn{2}{c}{Relative to independent training} \\
\cmidrule(lr){2-3} \cmidrule(lr){4-5}
Backbone & Wall time & Peak memory (GiB) & Train time & Peak memory \\
\midrule
Llama 3.2-1B & 3:19:46 & 16.92 & 0.840--0.964 & 1.000 \\
Llama 3.2-3B & 6:07:00 & 33.54 & 0.960--1.128 & 1.000--1.001 \\
Qwen2-1.5B-Instruct & 4:41:26 & 26.16 & 0.950--1.137 & 1.000 \\
Qwen3-4B & 8:37:52 & 48.44 & 1.005--1.012 & 1.000 \\
\bottomrule
\end{tabular}
\end{table}

\Cref{tab:cost-particles} shows how the cost depends on the number of particles.
Training takes between $2.94$ and $3.07$ GPU hours per particle on Llama 3.2-3B and between $4.23$ and $4.36$ on Qwen3-4B, and evaluation takes between $17.3$ and $17.8$ ms and between $24.8$ and $26.4$ ms per question per particle, so both costs grow linearly in $m$, to within $7\%$.
With $m=4$ each GPU holds the frozen backbone and a single particle, which is the configuration of a single adapter: $22.10$ GiB and $2.94$ GPU hours on Llama 3.2-3B, and $30.45$ GiB and $4.23$ GPU hours on Qwen3-4B.
At evaluation the particles are applied one at a time to the shared frozen backbone, so each additional particle adds only $63$ MiB of device memory on Llama 3.2-3B and $85$ MiB on Qwen3-4B.
Evaluation costs the same for every $\beta$: with $m=8$ it takes $71.6$, $141.3$, $89.0$ and $201.2$ ms per question for Llama 3.2-1B, Llama 3.2-3B, Qwen2-1.5B-Instruct and Qwen3-4B at $\beta=1$, and between $0.93$ and $1.06$ times as long for the other values of $\beta$.

\begin{table}[!tbp]
\centering
\caption{Cost against the number of particles $m$, for the bridge with $\beta=0.1$ and three epochs of training. GPU hours are the wall time of the training job times the number of GPUs, and peak memory is per GPU, each of which holds the frozen backbone and one particle for $m=4$ or two particles for $m\in\{8,16\}$. Evaluation time is per test question on one GPU, for all $m$ particles, at batch size $8$.}
\label{tab:cost-particles}
\small
\setlength{\tabcolsep}{4pt}
\begin{tabular}{@{}lrrccccc@{}}
\toprule
& & & \multicolumn{2}{c}{Training (GPU hours)} & Peak memory & \multicolumn{2}{c}{Evaluation (ms)} \\
\cmidrule(lr){4-5} \cmidrule(lr){7-8}
Backbone & $m$ & GPUs & Total & Per particle & (GiB) & Total & Per particle \\
\midrule
Llama 3.2-3B & 4 & 4 & 11.7 & 2.94 & 22.10 & 71.3 & 17.8 \\
 & 8 & 4 & 24.6 & 3.07 & 33.54 & 140.1 & 17.5 \\
 & 16 & 8 & 47.5 & 2.97 & 34.04 & 277.4 & 17.3 \\
\midrule
Qwen3-4B & 4 & 4 & 16.9 & 4.23 & 30.45 & 105.5 & 26.4 \\
 & 8 & 4 & 34.7 & 4.34 & 48.44 & 206.0 & 25.7 \\
 & 16 & 8 & 69.8 & 4.36 & 48.95 & 396.7 & 24.8 \\
\bottomrule
\end{tabular}
\end{table}

\subsection{Implementation of the Particle Schemes}
\label{app:mfld-derivation}

Both case studies update each particle $\bm{\theta}_j$ using the gradient $\nabla_{\bm{\theta}_j} V$ of the potential $V$ in \eqref{eq:intro-noisy-adamw}, with the mixture risk $R_{\mathrm{mix}}$ replaced by the bridge $R_\beta$; Case Study~1 uses \ac{MFLD} \eqref{eq:intro-mfld} and Case Study~2 uses noisy AdamW \eqref{eq:intro-noisy-adamw}.
From \eqref{eq:convex-responsibility},
\begin{align*}
\nabla_{\bm{\theta}_j} R_\beta(Q_m) = -\frac{1}{n}\sum_{i=1}^n \rho_{ij}^{(\beta)} \nabla_{\bm{\theta}} \log p_{\bm{\theta}_j}(y_i\mid x_i) ,
\end{align*}
so the gradient of the bridge differs from the gradient \eqref{eq:mix-gradient} of the mixture risk only in that the raw responsibilities $r_{ij}$ are replaced by the effective responsibilities $\rho_{ij}^{(\beta)} = (1-\beta)r_{ij}+\beta/m$.
The $r_{ij}$ and the per-particle gradients $\nabla_{\bm{\theta}}\log p_{\bm{\theta}_j}(y_i\mid x_i)$ are already computed when the mixture risk is used, so using the bridge in place of the mixture risk incurs a negligible additional computational cost.
Moreover, particles interact only through the scalars $w_{ij}$, so for a minibatch of size $B$ the particles need only exchange $Bm$ numbers per iteration.

\subsubsection{MFLD in Case Study 1}
\label{app:mfld-implementation}

The \ac{MFLD} \eqref{eq:intro-mfld}, with $R_\beta$ in place of $R_{\mathrm{mix}}$, was simulated using the Euler--Maruyama discretisation described in \Cref{sec:intro}, i.e. \eqref{eq:intro-noisy-adamw} with $\eta\bm{D}_j^k$ replaced by $\epsilon\nabla_{\bm{\theta}_j}V$ and noise standard deviation $\sqrt{2\epsilon}$, where the step size $\epsilon$ satisfies $\epsilon n=0.1$, i.e. $\epsilon=5\times10^{-4}$, $10^{-4}$ and $2\times10^{-5}$ for $n=200$, $1{,}000$ and $5{,}000$.
The first three quarters of each chain were discarded as burn-in, and diagnostics were averaged over the last quarter.
The following runs were performed, each with eight independent chains unless stated otherwise:
\begin{itemize}
\item \emph{Main comparison:} $m=32$, $n=1{,}000$ and $2\times10^{6}$ iterations ($t=200$), for all six values of $\beta$.
\item \emph{Number of particles:} $m=8$ with $2\times10^{6}$ iterations and $m=128$ with $5\times10^{5}$ iterations ($t=50$), for all six values of $\beta$; and $m=128$ with $2\times10^{6}$ iterations and four chains, for $\beta=0$ from all three initialisations and for $\beta>0$ from the prior.
\item \emph{Long chains:} $10^{7}$ iterations ($t=1{,}000$), for $\beta\in\{0,0.01\}$ with $m=32$ and for $\beta=0$ with $m=8$, continuing the chains above with the same seeds.
\item \emph{Sample size:} $n=200$ with $4\times10^{5}$ iterations ($t=200$) for all six values of $\beta$, and $n=5{,}000$ with $10^{6}$ iterations ($t=20$) for $\beta\in\{0,0.01,0.1\}$.
\item \emph{Step size:} the main comparison with half the step size and $4\times10^{5}$ iterations ($t=20$), for $\beta\in\{0,0.01,0.1\}$.
\end{itemize}
The particles were initialised either by independent draws from $\Pi$, or at a slot partition with intercept $+M$ for half of the particles and $-M$ for the other half, zero slopes and independent $\mathcal N(0,10^{-4})$ perturbations of all parameters, with $M\in\{8,16\}$.
A particle counts as steep along a direction when its slope vector $(\theta_1,\theta_2)$ has norm at least $3$ and lies within $30$ degrees of that direction; the counts of chains in the intended decomposition are the same for every norm threshold in $\{1,3,5\}$ and every angle in $\{15,30,45\}$ degrees.
The implementation was checked against finite differences of $V$, with relative error below $10^{-8}$, and at $\beta=1$, where the particles are independent, against a Laplace approximation of the Bayesian posterior, whose means it matches to within $0.006$ and whose standard deviations to within $1\%$.

\subsubsection{Noisy AdamW in Case Study 2}
\label{app:implemented-update}

Case Study~2 uses the noisy AdamW update \eqref{eq:intro-noisy-adamw} of \citet{nitanda2025propagation} and \citet{wang2026distributional}, applied to the training objective \eqref{eq:cs2-objective}.
The ridge penalty in \eqref{eq:cs2-objective} corresponds to the Gaussian prior $\Pi=\mathcal{N}(0,\lambda_0^{-1}\bm{I})$ in $V$, so the weight decay of AdamW was switched off.
The learning rate is $\eta=10^{-4}$, and with $n=170{,}300$ and $m=8$ the injected noise has standard deviation $\sqrt{2\eta/(mn)}\approx1.2\times10^{-5}$ per step; as explained in \Cref{sec:intro}, the trained particles are therefore interpreted as approximate minimisers of \eqref{eq:cs2-objective}, rather than as samples from $Q_\beta^{(n)}$.
The $\alpha$-R\'{e}nyi baseline of \Cref{subsec:exp-renyi} was trained with the same update, with $R_\beta$ replaced by the $\alpha$-R\'{e}nyi objective; the variant reported as \emph{without noise} omits the injected Gaussian noise, as in \citet[Algorithm~1]{cordero2026soft}.

\section{Additional Empirical Results}
\label{app:empirical}

This appendix reports additional results for Case Study~1 (\Cref{app: additional 1}) and Case Study~2 (\Cref{app:full-results-backbones,app:checkpoint-sensitivity,app:additional-mechanism,app:competence,app:sequence-space-full}), in the order in which they are referred to in \Cref{sec: experiments}.

\subsection{Case Study 1}
\label{app: additional 1}

\Cref{tab:cs1} reports, for each value of $\beta$ considered in Case Study~1, the held-out \ac{NLL} of the mixture predictive from each of the three initialisations, together with the calibration, responsibility and convergence diagnostics discussed in \Cref{subsec:cs1-numerics}.
Halving the step size changed the held-out \ac{NLL} at $t=20$ by at most $0.004$ at $\beta=0$, which is less than the spread across chains, and by less than $0.001$ for $\beta\in\{0.01,0.1\}$.
With $n=200$, the mean held-out \ac{NLL} of the three initialisations at $\beta=0$ lies between $0.390$ and $0.397$, against $0.395$ at $\beta=0.01$, while individual chains at $\beta=0$ started from the prior range from $0.382$ to $0.413$.

\begin{table}[!tbp]
\centering
\small
\caption{Case Study~1 with $m=32$, $n=1{,}000$ and $t=200$. Held-out \ac{NLL} of the mixture predictive for the three initialisations, mean $\pm$ sample standard deviation over eight chains of the average over $t$ from $150$ to $200$. 
Expected calibration error (ECE, ten equal-width bins) of the mixture predictive and the raw responsibility effective sample size (ESS) divided by $m$ are for the chains started from the prior. $t_{1/2}$ is the median \ac{MFLD} time, at a resolution of $0.05$, until the training-set gap $R_{\mathrm{mix}}-H(Y\mid X)$ has halved from the slot partition with $M=16$.}
\label{tab:cs1}
\setlength{\tabcolsep}{2.5pt}
\begin{tabular}{lcccccc}
\toprule
Method & Prior & Slot, $M=8$ & Slot, $M=16$ & ECE & ESS$/m$ & $t_{1/2}$ \\
\midrule
\ac{PrO} posterior, $\beta=0$ & $0.4086\pm0.0192$ & $0.3838\pm0.0140$ & $0.4130\pm0.0093$ & $0.063$ & $0.71$ & $13.50$ \\
Bridge, $\beta=0.01$ & $0.3764\pm0.0005$ & $0.3760\pm0.0003$ & $0.3762\pm0.0004$ & $0.030$ & $0.77$ & $1.08$ \\
Bridge, $\beta=0.1$ & $0.4254\pm0.0002$ & $0.4255\pm0.0001$ & $0.4254\pm0.0002$ & $0.054$ & $0.82$ & $0.15$ \\
Bridge, $\beta=0.2$ & $0.4605\pm0.0001$ & $0.4606\pm0.0001$ & $0.4606\pm0.0001$ & $0.052$ & $0.91$ & $0.10$ \\
Bridge, $\beta=0.3$ & $0.4736\pm{<}10^{-4}$ & $0.4736\pm{<}10^{-4}$ & $0.4736\pm0.0001$ & $0.046$ & $0.98$ & $0.10$ \\
Bayesian posterior, $\beta=1$ & $0.4752\pm{<}10^{-4}$ & $0.4752\pm{<}10^{-4}$ & $0.4752\pm{<}10^{-4}$ & $0.045$ & $1.00$ & $0.10$ \\
\bottomrule
\end{tabular}
\end{table}

\FloatBarrier

\subsection{Held-Out Results for All Backbones}
\label{app:full-results-backbones}

\begin{figure}[!tbp]
\centering
\includegraphics[width=\textwidth]{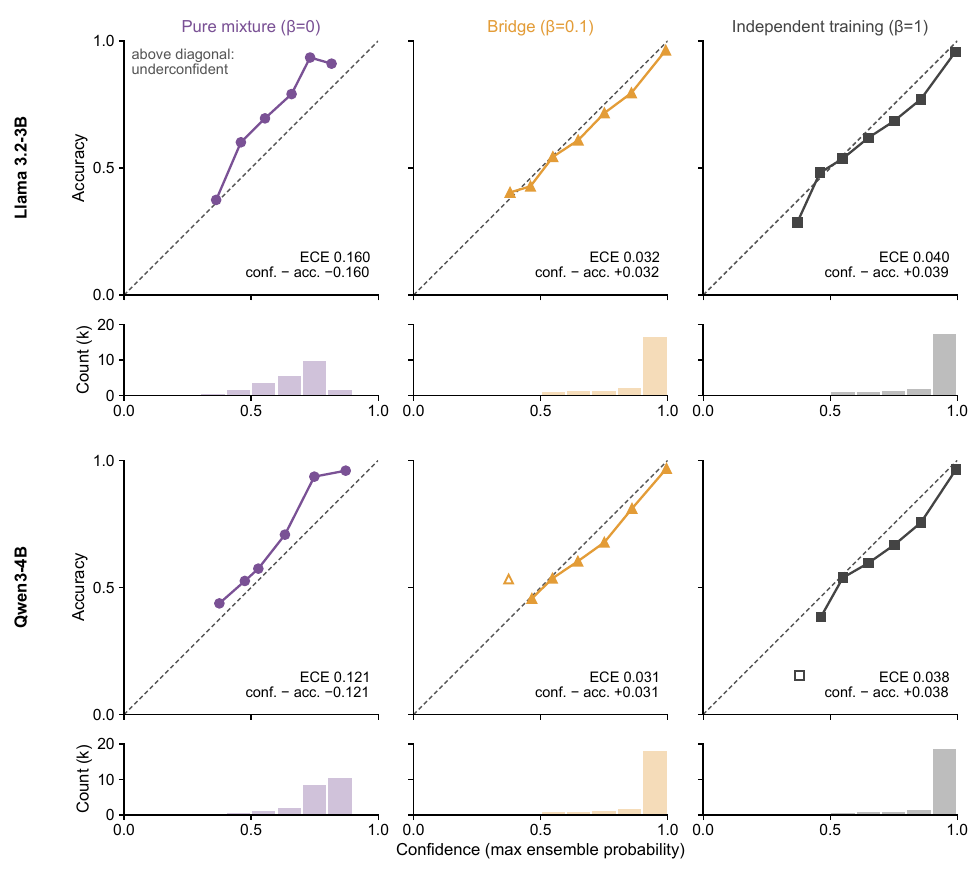}
\caption{Reliability diagrams for Llama 3.2-3B (top) and Qwen3-4B (bottom) after three epochs, pooled over the eight test sets ($22{,}419$ examples). Each point is one of ten equal-width confidence bins, where confidence is the largest probability of the mixture predictive. Points above the dashed diagonal are underconfident. Hollow markers are bins with fewer than $30$ examples, and bins with fewer than $5$ are not drawn. Strips below each panel give the number of examples per bin, in thousands.}
\label{fig:reliability}
\end{figure}

\Cref{tab:main-results} reports the held-out results for every backbone and value of $\beta$, averaged with equal weight over the eight test sets; these are the values plotted in \Cref{fig:beta-sweep}.
\Cref{tab:per-benchmark-accuracy,tab:per-benchmark-nll} report the accuracy and \ac{NLL} on each test set.
\Cref{fig:reliability} shows reliability diagrams for Llama 3.2-3B and Qwen3-4B: pure mixture is underconfident throughout, whereas with $\beta=0.1$ the curves follow the diagonal closely and the pooled ECE falls from $0.160$ to $0.032$ and from $0.121$ to $0.031$, respectively.
\Cref{tab:member-metrics} reports the held-out accuracy and \ac{NLL} of the individual particles.
Under pure mixture the particles are much weaker than their mixture: on Llama 3.2-3B the mean particle accuracy is $0.5961$ against $0.7832$ for the mixture, the worst particle attains $0.3320$, below the $0.3542$ of uniform guessing, and every particle has a higher \ac{NLL} than uniform guessing.
With $\beta=0.1$ every particle is accurate on Llama 3.2-3B, Qwen2-1.5B-Instruct and Qwen3-4B, the worst particle being within $0.031$ of the mixture in accuracy, and the mixture is more accurate and has a lower \ac{NLL} than each of its particles.

\begin{table}[!tbp]
\centering
\small
\setlength{\tabcolsep}{3.5pt}
\caption{Held-out results after three epochs, averaged with equal weight over the eight test sets. \ac{NLL} is the held-out $R_{\mathrm{mix}}$, and the signed gap is mean confidence minus accuracy. Bold marks the best value for each backbone (closest to zero for the signed gap).}
\label{tab:main-results}
\begin{tabular}{@{}llccccc@{}}
\toprule
Backbone & Setting & Acc.~$\uparrow$ & \ac{NLL}~$\downarrow$ & ECE~$\downarrow$ & Brier~$\downarrow$ & Signed gap \\
\midrule
Llama 3.2-1B & Pure mixture, $\beta=0$ & 0.3683 & 1.1162 & 0.0847 & 0.6591 & $0.0554$ \\
 & Bridge, $\beta=0.01$ & 0.3734 & 1.0861 & \textbf{0.0067} & 0.6417 & $\mathbf{-0.0013}$ \\
 & Bridge, $\beta=0.1$ & 0.6782 & 0.8584 & 0.1656 & 0.4950 & $-0.1633$ \\
 & Bridge, $\beta=0.2$ & 0.7076 & 0.8058 & 0.1702 & 0.4602 & $-0.1673$ \\
 & Bridge, $\beta=0.3$ & 0.3732 & 1.0851 & 0.0160 & 0.6407 & $0.0029$ \\
 & Independent training, $\beta=1$ & \textbf{0.7547} & \textbf{0.5915} & 0.0515 & \textbf{0.3369} & $0.0079$ \\
\midrule
Llama 3.2-3B & Pure mixture, $\beta=0$ & 0.7832 & 0.6104 & 0.1274 & 0.3354 & $-0.1195$ \\
 & Bridge, $\beta=0.01$ & 0.8454 & 0.4616 & \textbf{0.0496} & 0.2301 & $\mathbf{0.0441}$ \\
 & Bridge, $\beta=0.1$ & 0.8558 & \textbf{0.4271} & 0.0508 & 0.2163 & $0.0461$ \\
 & Bridge, $\beta=0.2$ & 0.8553 & 0.4277 & 0.0541 & 0.2156 & $0.0502$ \\
 & Bridge, $\beta=0.3$ & 0.8580 & 0.4395 & 0.0556 & \textbf{0.2141} & $0.0508$ \\
 & Independent training, $\beta=1$ & \textbf{0.8598} & 0.4513 & 0.0589 & 0.2143 & $0.0546$ \\
\midrule
Qwen2-1.5B-Instruct & Pure mixture, $\beta=0$ & 0.7962 & 0.6001 & \textbf{0.0631} & 0.2948 & $\mathbf{0.0599}$ \\
 & Bridge, $\beta=0.01$ & 0.8034 & 0.5987 & 0.0707 & 0.2886 & $0.0669$ \\
 & Bridge, $\beta=0.1$ & \textbf{0.8138} & \textbf{0.5510} & 0.0716 & 0.2773 & $0.0705$ \\
 & Bridge, $\beta=0.2$ & 0.8127 & 0.5524 & 0.0737 & \textbf{0.2752} & $0.0731$ \\
 & Bridge, $\beta=0.3$ & 0.8119 & 0.5597 & 0.0791 & 0.2773 & $0.0778$ \\
 & Independent training, $\beta=1$ & 0.8126 & 0.6030 & 0.0867 & 0.2810 & $0.0854$ \\
\midrule
Qwen3-4B & Pure mixture, $\beta=0$ & 0.8802 & 0.4059 & 0.1080 & 0.2046 & $-0.1018$ \\
 & Bridge, $\beta=0.01$ & 0.8880 & 0.3214 & 0.0570 & 0.1683 & $-0.0320$ \\
 & Bridge, $\beta=0.1$ & \textbf{0.8958} & 0.3185 & 0.0458 & \textbf{0.1588} & $0.0379$ \\
 & Bridge, $\beta=0.2$ & 0.8927 & 0.3212 & \textbf{0.0450} & 0.1613 & $0.0425$ \\
 & Bridge, $\beta=0.3$ & 0.8926 & \textbf{0.3099} & 0.0630 & 0.1608 & $\mathbf{-0.0264}$ \\
 & Independent training, $\beta=1$ & 0.8956 & 0.3334 & 0.0484 & 0.1602 & $0.0454$ \\
\bottomrule
\end{tabular}
\end{table}

\begin{table}[!tbp]
\centering
\caption{Held-out accuracy on each test set after three epochs. Macro is the average over the eight test sets with equal weight. Bold marks the best value in each column for each backbone. ARC-C and ARC-E denote ARC Challenge and ARC Easy, HellaS.\ HellaSwag, OBQA OpenBookQA, SIQA Social IQA and WinoG.\ WinoGrande.}
\label{tab:per-benchmark-accuracy}
\footnotesize
\setlength{\tabcolsep}{3pt}
\begin{tabular}{@{}lccccccccc@{}}
\toprule
Method & ARC-C & ARC-E & BoolQ & HellaS. & OBQA & PIQA & SIQA & WinoG. & Macro \\
\midrule
\multicolumn{10}{@{}l}{\emph{Llama 3.2-1B}} \\
Pure mixture, $\beta=0$ & 0.2270 & 0.2517 & 0.6217 & 0.2504 & 0.2760 & 0.4951 & 0.3291 & 0.4957 & 0.3683 \\
Bridge, $\beta=0.01$ & 0.2543 & 0.2542 & 0.6217 & 0.2474 & 0.2740 & 0.5125 & 0.3214 & 0.5020 & 0.3734 \\
Bridge, $\beta=0.1$ & 0.5162 & 0.7311 & 0.6465 & 0.7594 & 0.6380 & 0.7688 & 0.6965 & 0.6693 & 0.6782 \\
Bridge, $\beta=0.2$ & 0.5717 & 0.7479 & 0.6599 & 0.8160 & 0.6840 & 0.7775 & 0.7226 & 0.6811 & 0.7076 \\
Bridge, $\beta=0.3$ & 0.2551 & 0.2428 & 0.6217 & 0.2421 & 0.2820 & 0.4869 & 0.3460 & 0.5091 & 0.3732 \\
Independent training, $\beta=1$ & \textbf{0.6323} & \textbf{0.7988} & \textbf{0.6731} & \textbf{0.8815} & \textbf{0.7380} & \textbf{0.8156} & \textbf{0.7549} & \textbf{0.7435} & \textbf{0.7547} \\
\addlinespace[3pt]
\multicolumn{10}{@{}l}{\emph{Llama 3.2-3B}} \\
Pure mixture, $\beta=0$ & 0.7022 & 0.8636 & 0.6278 & 0.9109 & 0.7860 & 0.8270 & 0.7733 & 0.7751 & 0.7832 \\
Bridge, $\beta=0.01$ & 0.7730 & 0.9045 & 0.7453 & 0.9472 & 0.8500 & 0.8683 & 0.8142 & 0.8603 & 0.8454 \\
Bridge, $\beta=0.1$ & 0.7901 & \textbf{0.9188} & \textbf{0.7572} & 0.9551 & 0.8620 & 0.8857 & 0.8153 & 0.8619 & 0.8558 \\
Bridge, $\beta=0.2$ & 0.7961 & 0.9036 & 0.7523 & 0.9568 & 0.8620 & 0.8819 & 0.8183 & 0.8713 & 0.8553 \\
Bridge, $\beta=0.3$ & 0.7986 & 0.9146 & 0.7498 & 0.9564 & \textbf{0.8660} & 0.8868 & 0.8173 & 0.8745 & 0.8580 \\
Independent training, $\beta=1$ & \textbf{0.8046} & 0.9078 & 0.7557 & \textbf{0.9577} & \textbf{0.8660} & \textbf{0.8885} & \textbf{0.8193} & \textbf{0.8785} & \textbf{0.8598} \\
\addlinespace[3pt]
\multicolumn{10}{@{}l}{\emph{Qwen2-1.5B-Instruct}} \\
Pure mixture, $\beta=0$ & 0.7227 & 0.8674 & 0.6908 & 0.9046 & 0.8040 & 0.8156 & 0.7876 & 0.7766 & 0.7962 \\
Bridge, $\beta=0.01$ & 0.7287 & 0.8758 & 0.7043 & 0.9106 & 0.8160 & 0.8264 & 0.7820 & 0.7837 & 0.8034 \\
Bridge, $\beta=0.1$ & 0.7406 & 0.8733 & \textbf{0.7104} & 0.9216 & \textbf{0.8280} & 0.8346 & \textbf{0.7932} & 0.8082 & \textbf{0.8138} \\
Bridge, $\beta=0.2$ & 0.7355 & 0.8771 & 0.7083 & \textbf{0.9240} & 0.8180 & \textbf{0.8384} & 0.7907 & 0.8098 & 0.8127 \\
Bridge, $\beta=0.3$ & \textbf{0.7423} & 0.8733 & 0.7098 & 0.9231 & 0.8220 & 0.8264 & 0.7845 & \textbf{0.8137} & 0.8119 \\
Independent training, $\beta=1$ & 0.7312 & \textbf{0.8775} & 0.7067 & 0.9217 & \textbf{0.8280} & 0.8368 & 0.7851 & \textbf{0.8137} & 0.8126 \\
\addlinespace[3pt]
\multicolumn{10}{@{}l}{\emph{Qwen3-4B}} \\
Pure mixture, $\beta=0$ & 0.9053 & \textbf{0.9705} & 0.7401 & 0.9557 & 0.9220 & 0.8803 & 0.8066 & 0.8611 & 0.8802 \\
Bridge, $\beta=0.01$ & 0.9078 & 0.9680 & 0.7560 & 0.9613 & 0.9280 & 0.8966 & 0.8106 & 0.8753 & 0.8880 \\
Bridge, $\beta=0.1$ & \textbf{0.9121} & 0.9697 & \textbf{0.7587} & 0.9640 & 0.9360 & 0.9032 & \textbf{0.8234} & \textbf{0.8990} & \textbf{0.8958} \\
Bridge, $\beta=0.2$ & 0.9096 & 0.9676 & 0.7541 & 0.9639 & 0.9320 & 0.9075 & 0.8193 & 0.8879 & 0.8927 \\
Bridge, $\beta=0.3$ & 0.9104 & 0.9672 & 0.7529 & \textbf{0.9650} & \textbf{0.9400} & 0.9053 & 0.8193 & 0.8808 & 0.8926 \\
Independent training, $\beta=1$ & 0.9096 & 0.9697 & 0.7572 & 0.9647 & 0.9380 & \textbf{0.9108} & 0.8183 & 0.8966 & 0.8956 \\
\bottomrule
\end{tabular}
\end{table}

\begin{table}[!tbp]
\centering
\caption{Held-out \ac{NLL} on each test set after three epochs, with columns as in \Cref{tab:per-benchmark-accuracy}.}
\label{tab:per-benchmark-nll}
\footnotesize
\setlength{\tabcolsep}{3pt}
\begin{tabular}{@{}lccccccccc@{}}
\toprule
Method & ARC-C & ARC-E & BoolQ & HellaS. & OBQA & PIQA & SIQA & WinoG. & Macro \\
\midrule
\multicolumn{10}{@{}l}{\emph{Llama 3.2-1B}} \\
Pure mixture, $\beta=0$ & 1.4492 & 1.4412 & 0.6907 & 1.4309 & 1.4266 & 0.6933 & 1.1047 & 0.6932 & 1.1162 \\
Bridge, $\beta=0.01$ & 1.3860 & 1.3861 & 0.6598 & 1.3865 & 1.3858 & 0.6930 & 1.0987 & 0.6931 & 1.0861 \\
Bridge, $\beta=0.1$ & 1.1819 & 0.9742 & 0.6320 & 0.9515 & 1.0789 & 0.5618 & 0.8552 & 0.6318 & 0.8584 \\
Bridge, $\beta=0.2$ & 1.1203 & 0.9141 & 0.6224 & 0.8642 & 0.9851 & 0.5454 & 0.7834 & 0.6113 & 0.8058 \\
Bridge, $\beta=0.3$ & 1.3855 & 1.3864 & 0.6526 & 1.3871 & 1.3859 & 0.6930 & 1.0976 & 0.6924 & 1.0851 \\
Independent training, $\beta=1$ & \textbf{0.9411} & \textbf{0.5551} & \textbf{0.6004} & \textbf{0.3839} & \textbf{0.6960} & \textbf{0.4148} & \textbf{0.5929} & \textbf{0.5477} & \textbf{0.5915} \\
\addlinespace[3pt]
\multicolumn{10}{@{}l}{\emph{Llama 3.2-3B}} \\
Pure mixture, $\beta=0$ & 0.8574 & 0.5917 & 0.6389 & 0.5090 & 0.7184 & 0.4323 & 0.6345 & 0.5011 & 0.6104 \\
Bridge, $\beta=0.01$ & 0.6800 & 0.3127 & 0.6156 & 0.1714 & 0.5017 & 0.3540 & 0.5412 & 0.5164 & 0.4616 \\
Bridge, $\beta=0.1$ & 0.6724 & \textbf{0.2657} & \textbf{0.5883} & 0.1368 & 0.4451 & \textbf{0.3134} & 0.5165 & 0.4790 & \textbf{0.4271} \\
Bridge, $\beta=0.2$ & 0.6562 & 0.2830 & 0.6187 & \textbf{0.1338} & \textbf{0.4340} & 0.3199 & \textbf{0.5152} & \textbf{0.4611} & 0.4277 \\
Bridge, $\beta=0.3$ & \textbf{0.6532} & 0.2790 & 0.6250 & 0.1388 & 0.4654 & 0.3393 & 0.5280 & 0.4871 & 0.4395 \\
Independent training, $\beta=1$ & 0.7046 & 0.3084 & 0.6134 & 0.1447 & 0.4497 & 0.3475 & 0.5613 & 0.4809 & 0.4513 \\
\addlinespace[3pt]
\multicolumn{10}{@{}l}{\emph{Qwen2-1.5B-Instruct}} \\
Pure mixture, $\beta=0$ & 0.8736 & 0.4109 & \textbf{0.6122} & 0.3023 & 0.6320 & 0.5155 & 0.6356 & 0.8190 & 0.6001 \\
Bridge, $\beta=0.01$ & 0.8832 & 0.4030 & 0.6977 & 0.2862 & 0.6645 & 0.4772 & 0.6439 & 0.7339 & 0.5987 \\
Bridge, $\beta=0.1$ & \textbf{0.8116} & 0.3681 & 0.7327 & 0.2345 & \textbf{0.5394} & \textbf{0.4419} & \textbf{0.6196} & 0.6604 & \textbf{0.5510} \\
Bridge, $\beta=0.2$ & 0.8383 & \textbf{0.3674} & 0.7221 & \textbf{0.2338} & 0.5480 & 0.4432 & 0.6264 & 0.6402 & 0.5524 \\
Bridge, $\beta=0.3$ & 0.8292 & 0.3963 & 0.7226 & 0.2408 & 0.5569 & 0.4650 & 0.6467 & \textbf{0.6200} & 0.5597 \\
Independent training, $\beta=1$ & 0.9447 & 0.4150 & 0.7060 & 0.2659 & 0.5843 & 0.5296 & 0.6772 & 0.7016 & 0.6030 \\
\addlinespace[3pt]
\multicolumn{10}{@{}l}{\emph{Qwen3-4B}} \\
Pure mixture, $\beta=0$ & 0.4530 & 0.2810 & 0.5383 & 0.3214 & 0.4326 & 0.3126 & 0.5526 & 0.3561 & 0.4059 \\
Bridge, $\beta=0.01$ & 0.3190 & 0.1732 & \textbf{0.5227} & 0.1977 & 0.2957 & 0.2731 & \textbf{0.4764} & 0.3133 & 0.3214 \\
Bridge, $\beta=0.1$ & 0.3064 & \textbf{0.0917} & 0.6400 & 0.1132 & 0.2543 & 0.2632 & 0.5238 & 0.3556 & 0.3185 \\
Bridge, $\beta=0.2$ & 0.3288 & 0.1006 & 0.6376 & \textbf{0.1107} & \textbf{0.2267} & 0.2843 & 0.5425 & 0.3389 & 0.3212 \\
Bridge, $\beta=0.3$ & \textbf{0.3051} & 0.1694 & 0.5342 & 0.1852 & 0.2625 & \textbf{0.2525} & 0.4820 & \textbf{0.2881} & \textbf{0.3099} \\
Independent training, $\beta=1$ & 0.3405 & 0.1096 & 0.6388 & 0.1197 & 0.2299 & 0.3029 & 0.5580 & 0.3675 & 0.3334 \\
\bottomrule
\end{tabular}
\end{table}

\begin{table}[!tbp]
\centering
\caption{Held-out accuracy and \ac{NLL} of the individual particles after three epochs with $m=8$, next to those of their mixture (\Cref{tab:main-results}), averaged with equal weight over the eight test sets. The mean is over the eight particles, and the worst particle is the one with the lowest accuracy or the highest \ac{NLL}. Uniform guessing over the candidate answers has accuracy $0.3542$ and \ac{NLL} $1.0903$.}
\label{tab:member-metrics}
\footnotesize
\setlength{\tabcolsep}{3pt}
\begin{tabular}{@{}llcccccc@{}}
\toprule
& & \multicolumn{3}{c}{Accuracy $\uparrow$} & \multicolumn{3}{c}{\ac{NLL} $\downarrow$} \\
\cmidrule(lr){3-5} \cmidrule(lr){6-8}
& & & \multicolumn{2}{c}{Particles} & & \multicolumn{2}{c}{Particles} \\
\cmidrule(lr){4-5} \cmidrule(lr){7-8}
Backbone & Method & Mixture & Mean & Worst & Mixture & Mean & Worst \\
\midrule
Llama 3.2-1B & Pure mixture, $\beta=0$ & 0.3683 & 0.3535 & 0.3331 & 1.1162 & 12.7358 & 18.1907 \\
 & Bridge, $\beta=0.01$ & 0.3734 & 0.3576 & 0.3416 & 1.0861 & 1.1053 & 1.1165 \\
 & Bridge, $\beta=0.1$ & 0.6782 & 0.4674 & 0.3599 & 0.8584 & 1.1304 & 1.2423 \\
 & Bridge, $\beta=0.2$ & 0.7076 & 0.4956 & 0.3664 & 0.8058 & 1.0685 & 1.1141 \\
 & Bridge, $\beta=0.3$ & 0.3732 & 0.3698 & 0.3613 & 1.0851 & 1.0882 & 1.0987 \\
 & Independent training, $\beta=1$ & 0.7547 & 0.6814 & 0.3723 & 0.5915 & 0.9648 & 1.0866 \\
\midrule
Llama 3.2-3B & Pure mixture, $\beta=0$ & 0.7832 & 0.5961 & 0.3320 & 0.6104 & 5.3513 & 15.5145 \\
 & Bridge, $\beta=0.01$ & 0.8454 & 0.8118 & 0.8080 & 0.4616 & 0.9646 & 1.0104 \\
 & Bridge, $\beta=0.1$ & 0.8558 & 0.8287 & 0.8261 & 0.4271 & 0.7445 & 0.7838 \\
 & Bridge, $\beta=0.2$ & 0.8553 & 0.8314 & 0.8282 & 0.4277 & 0.7471 & 0.8010 \\
 & Bridge, $\beta=0.3$ & 0.8580 & 0.8347 & 0.8303 & 0.4395 & 0.7406 & 0.7515 \\
 & Independent training, $\beta=1$ & 0.8598 & 0.8390 & 0.8356 & 0.4513 & 0.7403 & 0.7587 \\
\midrule
Qwen2-1.5B-Instruct & Pure mixture, $\beta=0$ & 0.7962 & 0.7570 & 0.7467 & 0.6001 & 1.6077 & 2.5962 \\
 & Bridge, $\beta=0.01$ & 0.8034 & 0.7697 & 0.7626 & 0.5987 & 1.2392 & 1.2960 \\
 & Bridge, $\beta=0.1$ & 0.8138 & 0.7858 & 0.7830 & 0.5510 & 0.9458 & 0.9659 \\
 & Bridge, $\beta=0.2$ & 0.8127 & 0.7884 & 0.7850 & 0.5524 & 0.9353 & 0.9555 \\
 & Bridge, $\beta=0.3$ & 0.8119 & 0.7897 & 0.7868 & 0.5597 & 0.9407 & 0.9546 \\
 & Independent training, $\beta=1$ & 0.8126 & 0.7944 & 0.7886 & 0.6030 & 0.9482 & 0.9810 \\
\midrule
Qwen3-4B & Pure mixture, $\beta=0$ & 0.8802 & 0.7310 & 0.3417 & 0.4059 & 4.9112 & 17.7396 \\
 & Bridge, $\beta=0.01$ & 0.8880 & 0.8050 & 0.3682 & 0.3214 & 0.7382 & 1.0903 \\
 & Bridge, $\beta=0.1$ & 0.8958 & 0.8767 & 0.8716 & 0.3185 & 0.5577 & 0.5898 \\
 & Bridge, $\beta=0.2$ & 0.8927 & 0.8771 & 0.8749 & 0.3212 & 0.5490 & 0.5647 \\
 & Bridge, $\beta=0.3$ & 0.8926 & 0.8173 & 0.3847 & 0.3099 & 0.6109 & 1.0976 \\
 & Independent training, $\beta=1$ & 0.8956 & 0.8823 & 0.8808 & 0.3334 & 0.5224 & 0.5439 \\
\bottomrule
\end{tabular}
\end{table}

\paragraph{Training-Set Audit}
\label{app:training-audit}
To explain the entries of \Cref{tab:pro_optimisation_puzzle}, we examined the training-set \ac{NLL} of the individual particles of Qwen3-4B.
Under pure mixture, two of the eight particles have a mean training \ac{NLL} of about $17$ and $13$ nats.
At $\beta=0.3$ one particle fits the training set poorly, with a mean \ac{NLL} of about $0.96$ against at most $0.02$ for the other seven particles, which is why $\beta=0.3$ attains a higher training-set $R_{\mathrm{mix}}$ than $\beta=0.1$ and $\beta=0.2$.
At $\beta=0.01$ one particle reaches about $0.97$, and the other seven, at $0.27$ to $0.37$, also fit the training set about five times worse than at $\beta=0.1$.
At $\beta\in\{0.1,0.2,1\}$ the eight particles fit the training set almost equally well.

\FloatBarrier

\subsection{Robustness to Checkpoint Selection}
\label{app:checkpoint-sensitivity}

\begin{figure*}[!tbp]
\centering
\includegraphics[width=\textwidth]{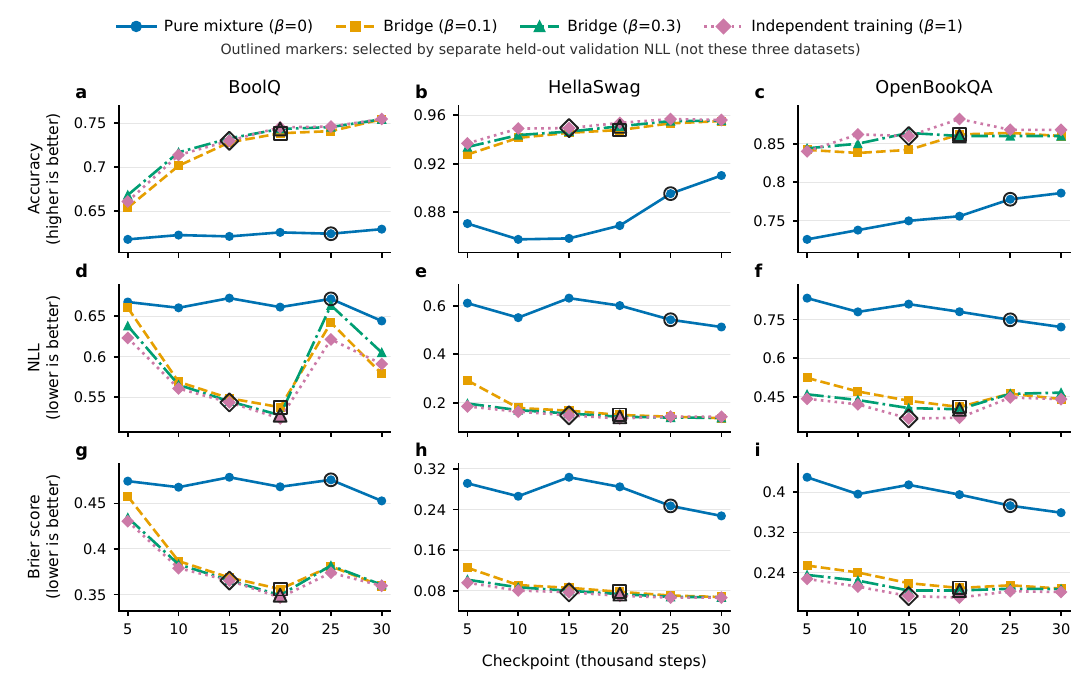}
\caption{Held-out performance of Llama 3.2-3B at each saved checkpoint on BoolQ, HellaSwag and OpenBookQA. Outlined markers indicate the checkpoint selected for each setting.}
\label{fig:llama3b_revised_checkpoint_predictive}
\end{figure*}

For Llama 3.2-3B and $\beta\in\{0,0.1,0.3,1\}$, we also selected, separately for each setting, the checkpoint with the lowest \ac{NLL} on a validation set of $120$ examples, from the checkpoints saved every $5{,}000$ steps.
The held-out results at the selected checkpoints are reported in \Cref{tab:llama3b_revised_selected_aggregate}, and the held-out performance at every saved checkpoint is shown in \Cref{fig:llama3b_revised_checkpoint_predictive}.
Pure mixture remains substantially worse than every positive value of $\beta$ in both accuracy and \ac{NLL}.
However, the ranking among the positive values of $\beta$ changes: at the final checkpoint the bridge with $\beta=0.1$ attains the lowest \ac{NLL}, whereas at the selected checkpoints independent training does.

\begin{table}[!tbp]
\centering
\caption{Held-out results for Llama 3.2-3B at the checkpoint selected for each setting by the lowest \ac{NLL} on a validation set of $120$ examples, averaged with equal weight over the eight test sets. Bold marks the best value in each column.}
\label{tab:llama3b_revised_selected_aggregate}
\small
\begin{tabular}{@{}lccccc@{}}
\toprule
Method & Checkpoint & Acc.~$\uparrow$ & \ac{NLL}~$\downarrow$ & ECE~$\downarrow$ & Brier~$\downarrow$ \\
\midrule
Pure mixture, $\beta=0$ & 25k & 0.7736 & 0.6335 & 0.1405 & 0.3515 \\
Bridge, $\beta=0.1$ & 20k & 0.8474 & 0.3902 & 0.0309 & 0.2201 \\
Bridge, $\beta=0.3$ & 20k & \textbf{0.8510} & 0.3842 & \textbf{0.0284} & \textbf{0.2139} \\
Independent training, $\beta=1$ & 15k & 0.8483 & \textbf{0.3804} & 0.0298 & 0.2156 \\
\bottomrule
\end{tabular}
\end{table}

\FloatBarrier

\subsection{Additional Mechanism Diagnostics}
\label{app:additional-mechanism}

\begin{figure*}[!tbp]
  \centering
  \begin{subfigure}[t]{0.485\textwidth}
    \vspace{0pt}
    \centering
    \includegraphics[width=\linewidth]{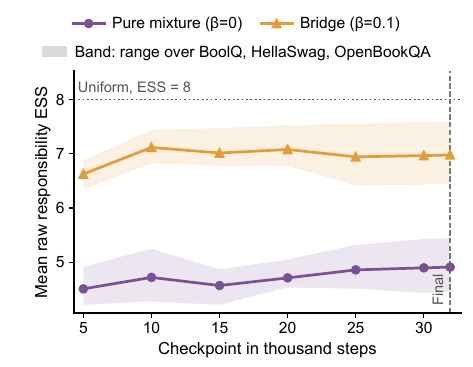}
    \caption{Raw responsibility ESS.}
  \end{subfigure}\hfill%
  \begin{subfigure}[t]{0.485\textwidth}
    \vspace{0pt}
    \centering
    \includegraphics[width=\linewidth]{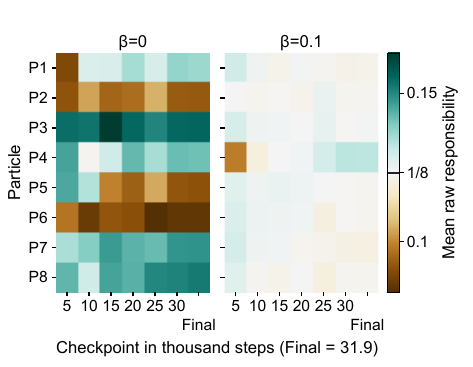}
    \caption{Mean raw responsibility per particle.}
  \end{subfigure}
  \caption{Further post hoc diagnostics for Llama 3.2-3B under pure mixture ($\beta=0$) and the bridge ($\beta=0.1$), complementing the top row of \Cref{fig:mechanism}. (a) Raw responsibility \ac{ESS}. (b) Mean raw responsibility of each particle.}
  \label{fig:starvation_dynamics_llama3b}
\end{figure*}

\begin{figure*}[!tbp]
  \centering
  \begin{subfigure}[t]{0.485\textwidth}
    \vspace{0pt}
    \centering
    \includegraphics[width=\linewidth]{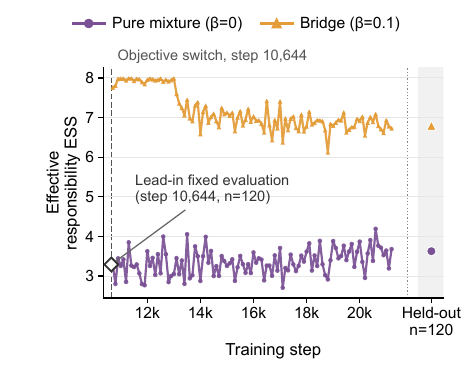}
    \caption{Effective responsibility ESS.}
  \end{subfigure}\hfill%
  \begin{subfigure}[t]{0.485\textwidth}
    \vspace{0pt}
    \centering
    \includegraphics[width=\linewidth]{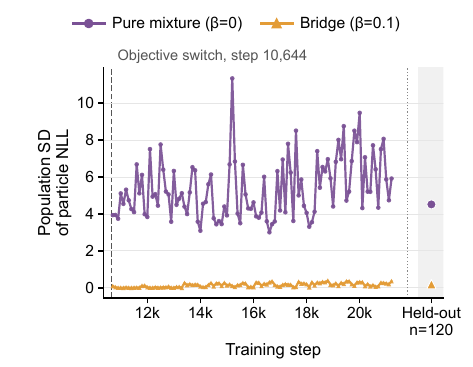}
    \caption{Particle-level \ac{NLL} dispersion.}
  \end{subfigure}
  \caption{Further diagnostics for the objective-switch experiment for Llama 3.2-3B, complementing the bottom row of \Cref{fig:mechanism}: training is continued from the pure-mixture particles after one epoch with $\beta=0$ or $\beta=0.1$. (a) \ac{ESS} of the effective responsibilities. (b) Standard deviation across particles of the particle-level \ac{NLL}. Values at the right of each panel are computed on a held-out set of $120$ examples.}
  \label{fig:objective_switch_rescue}
\end{figure*}

\begin{figure*}[!tbp]
  \centering
  \begin{subfigure}[t]{0.485\textwidth}
    \centering
    \includegraphics[width=\linewidth]{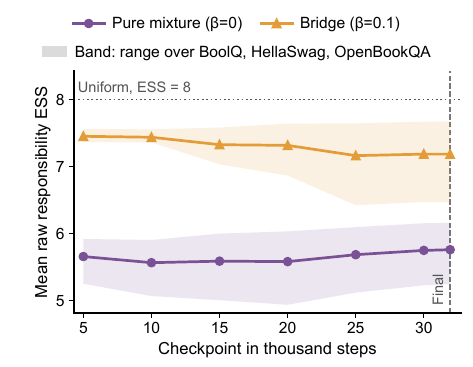}
    \caption{Raw responsibility ESS.}
  \end{subfigure}\hfill
  \begin{subfigure}[t]{0.485\textwidth}
    \centering
    \includegraphics[width=\linewidth]{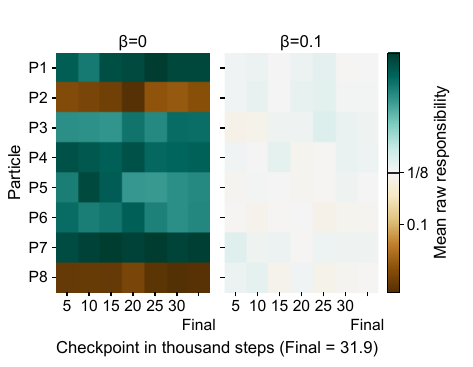}
    \caption{Mean raw responsibility per particle.}
  \end{subfigure}
  \vspace{0.6em}
  \begin{subfigure}[t]{0.485\textwidth}
    \centering
    \includegraphics[width=\linewidth]{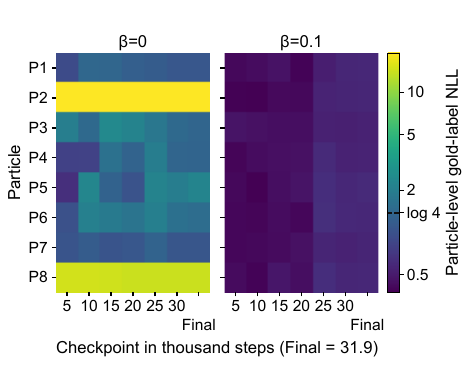}
    \caption{Particle-level gold-label \ac{NLL}.}
  \end{subfigure}\hfill
  \begin{subfigure}[t]{0.485\textwidth}
    \centering
    \includegraphics[width=\linewidth]{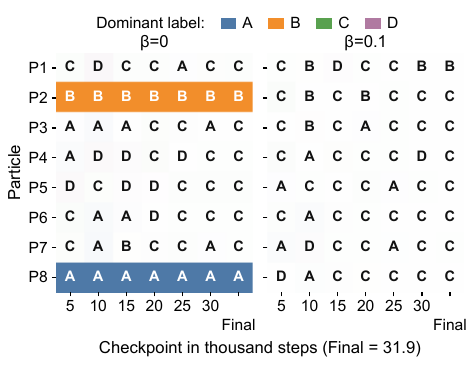}
    \caption{Label occupancy on HellaSwag.}
  \end{subfigure}
  \caption{Post hoc diagnostics for Qwen3-4B under pure mixture ($\beta=0$) and the bridge ($\beta=0.1$), corresponding to \Cref{fig:starvation_dynamics_llama3b} and the top row of \Cref{fig:mechanism} for Llama 3.2-3B. (a) Raw responsibility \ac{ESS}. (b) Mean raw responsibility of each particle. (c) Particle-level \ac{NLL}. (d) Most frequent label of each particle on HellaSwag.}
  \label{fig:qwen3_starvation_dynamics}
\end{figure*}

\Cref{fig:starvation_dynamics_llama3b,fig:objective_switch_rescue} report the remaining diagnostics for Llama 3.2-3B, for the post hoc analysis and the objective-switch experiment respectively.
\Cref{fig:qwen3_starvation_dynamics} reports the corresponding post hoc diagnostics for Qwen3-4B, and \Cref{tab:m_ablation} gives the full results of the experiment varying the number of particles, summarised in \Cref{fig:m-ablation}.

\begin{table}[!tbp]
\centering
\caption{Varying the number of particles $m$, with held-out results after three epochs averaged with equal weight over the eight test sets; the rows with $m=8$ are the main runs. ESS$/m$ is the raw responsibility \ac{ESS} divided by $m$, and $\Pr(r<0.01)$ is the fraction of near-starved pairs of example and particle. Bold marks the best value over $m$ for each backbone and setting.}
\label{tab:m_ablation}
\small
\setlength{\tabcolsep}{4pt}
\begin{tabular}{@{}llrccccc@{}}
\toprule
Backbone & Method & $m$ & Acc.~$\uparrow$ & \ac{NLL}~$\downarrow$ & Signed gap & ESS$/m$ & $\Pr(r<0.01)$ \\
\midrule
Llama 3.2-3B & Pure mixture, $\beta=0$ & 4 & \textbf{0.8428} & \textbf{0.5700} & $\mathbf{0.0638}$ & 0.8657 & 0.0811 \\
 &  & 8 & 0.7832 & 0.6104 & $-0.1195$ & 0.6354 & 0.2946 \\
 &  & 16 & 0.6542 & 0.9596 & $-0.2251$ & 0.4047 & 0.5927 \\
\addlinespace[2pt]
 & Bridge, $\beta=0.1$ & 4 & 0.8545 & 0.5049 & $0.0629$ & 0.8888 & 0.0456 \\
 &  & 8 & 0.8558 & 0.4271 & $0.0461$ & 0.8694 & 0.0689 \\
 &  & 16 & \textbf{0.8599} & \textbf{0.3904} & $\mathbf{-0.0309}$ & 0.8212 & 0.1013 \\
\midrule
Qwen3-4B & Pure mixture, $\beta=0$ & 4 & \textbf{0.8877} & 0.4667 & $\mathbf{0.0495}$ & 0.9003 & 0.0659 \\
 &  & 8 & 0.8802 & \textbf{0.4059} & $-0.1018$ & 0.7371 & 0.2512 \\
 &  & 16 & 0.8535 & 0.7177 & $-0.3242$ & 0.5090 & 0.4846 \\
\addlinespace[2pt]
 & Bridge, $\beta=0.1$ & 4 & 0.8904 & 0.3811 & $0.0515$ & 0.9157 & 0.0356 \\
 &  & 8 & \textbf{0.8958} & \textbf{0.3185} & $\mathbf{0.0379}$ & 0.9054 & 0.0521 \\
 &  & 16 & 0.8937 & 0.3335 & $-0.0708$ & 0.8364 & 0.0725 \\
\bottomrule
\end{tabular}
\end{table}

\FloatBarrier

\subsection{Backbone Competence}
\label{app:competence}

\begin{figure}[!tbp]
\centering
\includegraphics[width=\textwidth]{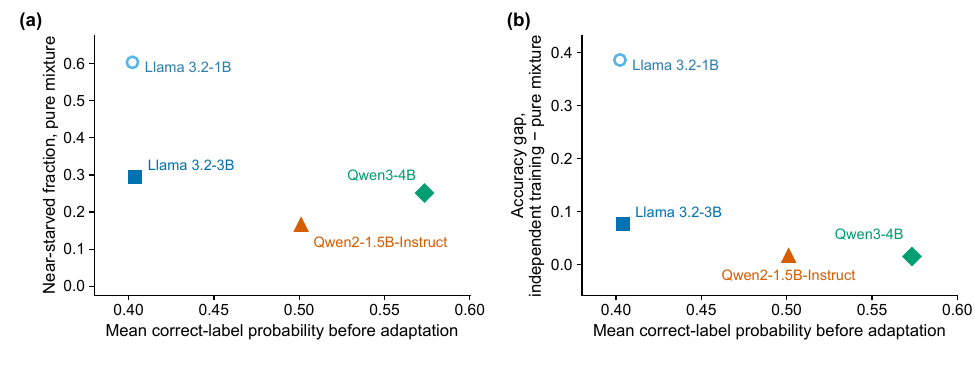}
\caption{Competence of the frozen backbone before fine-tuning, measured as the mean correct-label probability on the $170{,}300$ training examples, against (a) the fraction of near-starved pairs under pure mixture and (b) the accuracy of independent training minus that of pure mixture, both averaged over the eight test sets.}
\label{fig:competence}
\end{figure}

\Cref{tab:pre_adaptation} summarises the competence of each frozen backbone before fine-tuning, through the distribution over the training examples of its correct-label probability, which is the quantity that controls the implicit responsibility floor in \Cref{cor:competence-floor}.
\Cref{fig:competence} plots the mean of this distribution against two measures of the benefit of $\beta>0$, as discussed in \Cref{subsec:exp-competence}.

\begin{table}[!tbp]
\centering
\small
\caption{Competence of the frozen backbones before fine-tuning. Zero-shot accuracy is the macro accuracy on the eight test sets. The other columns summarise the distribution of the correct-label probability over the $170{,}300$ training examples, including the fraction of examples on which it is below $1/8$.}
\label{tab:pre_adaptation}
\begin{tabular}{lccccc}
\toprule
Backbone & Zero-shot acc. & Mean & Median & 10\% quantile & Below $1/8$ \\
\midrule
Llama 3.2-1B & 0.368 & 0.402 & 0.294 & 0.092 & 0.158 \\
Llama 3.2-3B & 0.370 & 0.404 & 0.223 & 0.076 & 0.266 \\
Qwen2-1.5B-Instruct & 0.570 & 0.501 & 0.469 & 0.122 & 0.102 \\
Qwen3-4B & 0.693 & 0.573 & 0.619 & 0.141 & 0.084 \\
\bottomrule
\end{tabular}
\end{table}

\FloatBarrier

\subsection{Implementation of the \texorpdfstring{$\alpha$}{alpha}-R\'enyi Comparison}
\label{app:sequence-space-full}

For \Cref{subsec:exp-renyi}, we instantiated the sequence-space $\alpha$-R\'enyi objective of \citet{cordero2026soft} on Llama 3.2-3B under our training set-up, with $m=8$ particles, three epochs and their principal setting $\alpha=0.8$, both with and without injected noise (\Cref{app:implemented-update}).
Their objective, and the predictive distribution that it optimises, average the likelihoods of complete answer sequences across particles before normalising over the candidate set, whereas the bridge uses the class-label predictive in which the answer-letter probabilities \eqref{eq:class-probability} are normalised within each particle before averaging.
In \Cref{tab:renyi-native-comparison} each method is therefore evaluated with the predictive distribution that it optimises.
Evaluating the $\alpha$-R\'enyi particles under the class-label predictive instead gives a macro accuracy of only $0.368$ (without noise) and $0.396$ (with noise), close to chance, so a comparison using a common predictive distribution would not be informative.